\documentclass[11pt,a4paper]{article}
\usepackage[utf8]{inputenc}
\usepackage[hyperref]{conll-2019}
\usepackage{times}
\usepackage{latexsym}
\usepackage{amsmath} 
\usepackage{float}
\usepackage{url}
\usepackage{graphicx}
\usepackage{multicol,multirow} 
\usepackage{tipa} 
\usepackage{subfig} 
\usepackage{placeins}

\aclfinalcopy

\title{Word Recognition, Competition, and Activation \\ in a Model of Visually Grounded Speech}

\author{
   {William N. Havard$^{1,2}$, Jean-Pierre Chevrot$^{2}$, Laurent Besacier$^{1}$}\\\\
   {$^{1}$ LIG, Univ. Grenoble Alpes, CNRS, Grenoble INP, 38000 Grenoble, France}\\
   {$^{2}$ LIDILEM, Univ. Grenoble Alpes, 38000 Grenoble, France}\\
   {\tt\small first-name.lastname@univ-grenoble-alpes.fr}
}

\begin{document}
\maketitle
\begin{abstract}
\label{doc:abstract}
In this paper, we study how word-like units are represented and activated in a recurrent neural model of visually grounded speech. The model used in our experiments is trained to project an image and its spoken description in a common representation space. We show that a recurrent model trained on spoken sentences implicitly segments its input into word-like units and reliably maps them to their correct visual referents. We introduce a methodology originating from linguistics to analyse the representation learned by neural networks -- the gating paradigm -- and show that the correct representation of a word is only activated if the network has access to first phoneme of the target word, suggesting that the network does not rely on a global acoustic pattern. Furthermore, we find out that not all speech frames (MFCC vectors in our case) play an equal role in the final encoded representation of a given word, but that some frames have a crucial effect on it. Finally, we suggest that word representation could be activated through a process of lexical competition.
\end{abstract}

\section{Introduction}
\label{doc:introduction}
Neural models of Visually Grounded Speech (VGS) sparked interest in linguists and cognitive scientists as they are able to incorporate multiple modalities in a single network and allow the analysis of complex interactions between them. Analysing these models does not only help to understand their technological limitations, but may also yield insight on the cognitive processes at work in humans \cite{DUPOUX18} who learn from contextually grounded speech utterances  (either visually, haptically, socially, etc.). This is with this idea in mind that one of the first computational model of visually grounded word acquisition was introduced by \citet{Roy02_CELL}. More recently, \citet{harwath_2016} and \citet{Chrupala2017} were among the first to propose neural models integrating these two modalities.

While \citet{Chrupala2017} and \citet{Alishahi2017} focused on analysing speech representations learnt by speech-image neural models from a phonological and semantic point of view, the present work focuses on lexical acquisition and the way speech utterances are segmented into lexical units by a neural model.

More precisely, we aim at understanding how word-like units are processed by a VGS architecture. First, we study if such models are robust to isolated word stimuli. As such networks are trained on raw speech utterances, robustness to isolated word stimuli would indicate that a segmentation process was implicitly carried out at training time. We also explore which factors influence the most such  \textit{word recognition}. In a second step, to better understand how individual words are activated by the network, we adapt the gating paradigm initially introduced to study human word recognition \cite{Grosjean1980} where our  neural model is inputted with speech segments of increasing duration (\textit{word activation}). Finally, as some linguistic
models assume that the first phoneme of a target word activates all the words starting by the same phoneme, we investigate if such a pattern holds true for our neural model as well (\textit{word competition}).  As far as we know, no other study has examined patterns of word recognition, activation and competition in models of VGS.

This paper is organised as follows: section \ref{doc:related-work} presents related works and section \ref{settings}  details our experimental material (data and model). Our contributions follow in section \ref{doc:word-recognition_intro} (word recognition), section \ref{doc:word-activation_intro} (word activation) and section \ref{doc:word-competition_intro} (word competition). Section~\ref{concl} concludes this work.

\section{Related Work}
\label{doc:related-work}
In this section we explore what is known about word recognition in humans. We then review recent works related to the representation of language in VGS models. A few words are also said about modified inputs and adversarial attacks as they are related to the analysis methodology used in part of this work.

\subsection{Word Recognition in Humans}
\label{doc:related-work_human-activation}
Many psycholinguistic models try to account for how words are activated and recognised from fluent speech. The process of word recognition \textit{``requires matching the spoken input with mental representations associated with word candidates"} \cite{dahan_magnuson}. One of the first model trying to account for how humans recognise and extract words from fluent speech is the \textsc{cohort} model by \citet{marslenwillow78_cohort}. In this model, word recognition proceeds in $3$ steps: \textit{access}, \textit{selection} and \textit{integration}. \textit{Access} denotes the process by which a set of words (a cohort) becomes activated if their onsets are consistent with the perceived spoken input. As soon as a word form becomes inconsistent with the spoken input, it is removed from the initial cohort (\textit{selection} phase). A word is deemed recognised as soon it is the last one standing in the cohort. \textit{Integration} consists in checking if the word's syntactic and semantic properties are consistent with the rest of the utterance. However, \textsc{cohort} supposes a full match between the perceived input and the word forms and does not account for word frequency in the access phase. \textsc{revised cohort} \cite{marslenwillow87_cohort} later relaxed the constraints on the cohort formation to take into account these facts. There is no active competition \textit{per se} between words in the \textsc{cohort} model. That is, the strength of activation of a word does not depend on the value of the activation of the other words, but only on how well the internalised word form matches the perceived spoken input. \textsc{trace} \cite{mcclelland_trace} is a connectionist model of spoken word recognition consisting of three layers of nodes, where each layer represents a particular linguistic unit (feature, phoneme and word). Layers are linked by exitatory connections (e.g. fricative feature node would activate /f/ phoneme node which would, in turn, activate words starting with this sound), and nodes within a layer are linked by inhibitory connections, thus inducing a real competition between activated words.
Contrary to the \textsc{cohort} model which does not allow words embedded in longer words to be activated, \textsc{trace} allows such activation. \textsc{shortlist} \cite{Norris_1994_Shortlist} is another model which builds upon \textsc{cohort} and \textsc{trace} by taking into consideration other features such as word stress.\footnote{For a review of spoken word recognition model, reader can consult \citet{dahan_magnuson} and \citet{weber_word-recogntion_review}}

To sum up, models of spoken word recognition consider that a set of words matching to a certain extent the spoken input is simultaneously activated and these models involve at some point a form of competition between the set of activated words before reaching the stage of recognition.

\subsection{Computational Models of VGS}
\label{doc:related-work_VGS-Models}
\citet{Roy02_CELL} were among the first to propose a computational model, known as CELL, that integrates both speech and vision to study child language acquisition. However, CELL required both speech and images to be pre-processed, where canonical shapes were first extracted from images and further represented as histograms; and speech was discretised into phonemes. More recently, CNN-based VGS models \cite{harwath_2016, Harwath18_words, Kamper_19} and RNN-based VGS models \cite{Chrupala2017} which do not require speech to be discretised into sub-units were introduced.
\citet{Chrupala2017} investigated how RNN-based models encode language, and showed such models tend to encode semantic information in higher layers, while form is better encoded in lower layers. \citet{Alishahi2017} studied if such models capture phonological information and showed that some layers do capture such information more accuratly than others. \citet{Kadar17_FormFunction} introduced \textit{omission scores} to interpret the contribution of individual tokens in text-based VGS models.
More recently, \citet{Havard2019} studied the behaviour of attention in RNN-based VGS models and showed that these models tend to focus on nouns and could display language-specific patterns, such as focusing on particules when prompted with Japanese.
Recently, \citet{Harwath18_words} showed that CNN-based models could reliably map word-like units to their visual referents, and \citet{Harwarth19_subword} showed such networks were sensitive to diphone transitions and that these were useful for the purpose of word recognition.
However, none of the aforementioned works studied the process by which words are recognised and activated. This present work aims at bridging what is known about word activation and recognition in humans and the computations at work in VGS models.

\subsection{Modified Inputs and Adversarial Attacks}

As will be shown later, the gating method used in this article modifies the input stimulus to better understand the behaviour of the neural model.

We can draw a parallel with approaches  recently introduced to show the vulnerability of deep networks to strategically modified samples (adversarial examples) and   to detect their over-sensitivity and over-stability points.
It was shown that imperceptible perturbations can fool the neural models to give false predictions. Inspired by the researches for images \cite{Su19_onePixel}, efforts on attacking neural networks for NLP applications emerged recently (see \citet{adversarial_survey} for a survey).
However, while a lot of references can be found for textual adversarial examples, fewer papers addressed adversarial attacks for speech (we can however mention the work of \citet{EURECOM+4436} addressing spoofing attacks in speaker verification and of \citet{DBLP:journals/corr/abs-1801-01944} attacking \textit{DeepSpeech} end-to-end ASR system).

\section{Experimental Settings}
\label{settings}

\subsection{Model Type}
Even though the methodologies developed in this work could also be applied to CNN-based VGS models, the present work will solely focus on the analysis of the representations learned by a RNN-based VGS model. Indeed, from a cognitive perspective, RNN-based models are more realistic than CNN-based models as the speech signal -- or in our case, a sequence of MFCC vectors -- is sequentially processed from left-to-right, whereas in CNN-based models the network processes multiple frames at the same time. This will thus allow us to explore if RNN-based models display human-like behaviour or not.

\subsection{Model Architecture}
\label{doc:experimental-settings_model-architecture}
The model we use for our experiment is based on that of \citet{Chrupala2017} and later modified by \citet{Havard2019}. It is trained to solve an image retrieval task: given a speech query, the model should retrieve the closest matching image. The model consists of two parts: an image encoder and a speech encoder. The image encoder takes VGG-16 pre-calculated vectors as input instead of raw images. It consists of a dense layer which reduces the 4096 dimensional VGG-16 input vector into a 512 dimensional vector which is then L2 normalised. The speech encoder takes 13 Mel Frequency Cepstral Coefficients (MFCC) vectors instead of raw speech.\footnote{12 mel frequency cepstral coefficients + log of total frame energy, vectors extracted every 10ms on a 25ms window} It consists of a convolutional layer (64 filters of length 6 and stride 3) followed by 5 stacked unidirectional GRU layers \cite{GRU}, with 512 units each. Two attention mechanisms \cite{AttnBahdanau} are used: one after the 1$^{st}$ recurrent layer and one after the 5$^{th}$ recurrent layer. The final vector produced by the speech encoder corresponds to the dot product of the weighted vectors outputted by each attention mechanism.
The model is trained to minimise the following triplet loss function as implemented by \citet{Chrupala2017}:
%
%

\begin{equation}
  \resizebox{1\hsize}{!}{
  \begin{math}
    \begin{split}
    \mathcal{L}(u, i, \alpha)=
      \sum_{u, i} 
      \Bigg( \sum_{u'}\max [0, \alpha + d(u, i) - d(u', i)]\\
      +\sum_{i'}\max[0, \alpha + d(u, i) - d(u, i')] \Bigg)
    \end{split}
  \end{math}
  }
  \label{eq:loss_fct}
\end{equation}

%
%

The loss function encourages the network to minimise the cosine distance $d$ between an image $i$ and its corresponding spoken description $u$ by a given margin $\alpha$ while maximising the distance between mismatching image/utterance pairs. For our experiments, we set $\alpha=0.2$.

\subsection{Data}
\label{doc:experimental-settings_data}
The data set used for our experiments is based on MSCOCO \cite{MSCOCO}. MSCOCO is a data set used to train computer vision systems, and features annotated images, each paired with 5 human written descriptions in English. MSCOCO's images where selected so that the images would contain instances of 80 possible object categories. We trained our model on the spoken extension introduced by \citet{SynthCOCO}. This extension  provides spoken version of the human written captions. It is worth mentioning that this extended data set features synthetic speech (female voice generated using Google's Text-To-Speech (TTS) system) and not real human speech.

\subsection{Model Training and Results}
We trained our model for 15 epochs with Adam optimiser and an initial learning rate of 0.0002. The training set comprises 113,287 images with 5 spoken captions per image. Validation and test set comprise 5000 images each.\footnote{The train/dev/test correspond to those used in \cite{Chrupala2017}} Model is evaluated in term of Recall@k (R@k) and median rank $\widetilde{r}$. That is, given a spoken query, which corresponds to a full utterance, we evaluate the model's ability to rank the unique paired image in the top $k$ images. We obtain a $\widetilde{r}$ of $28$.

Full results are shown in Table \ref{tab:results}. Even though our results are lower than the original implementation by \citet{Chrupala2017}, our model still performs far above chance level, showing it did learn how to map an image and its spoken description.

\begin{table}[htbp]
  \centering
    \begin{tabular}{c|c|c|c|c}
      \textbf{Model} & \textbf{R@1} & \textbf{R@5} & \textbf{R@10} & \textbf{$\widetilde{r}$}\\\hline
      Synth. COCO  & 0.056 & 0.182 & 0.284 & 28\\
    \end{tabular}
  \caption{
    Recall at 1, 5, and 10 results and median rank \textbf{$\widetilde{r}$} on a speech-image retrieval task (test part of our datasets with 5k images). \citet{Chrupala2017} with RHN reports median rank \textbf{$\widetilde{r}=13$}. Chance median rank $\widetilde{r}$ is 2500.5.
    }
  \label{tab:results}
\end{table}

\section{Word Recognition}
\label{doc:word-recognition_intro}
\citet{Harwath18_words} observed that CNN-based models can reliably map word-like units to their corresponding visual reference. \citet{Chrupala2017} and more recenlty \citet{Merkx2019LanguageLU} showed that RNN-based utterance embeddings contain information about individual words, but did not show for what type of words this behaviour holds true and if the model had learnt to map these individual words to their visual referents. \citet{Havard2019} showed that the attention mechanism of RNN-based VGS models tends to focus on the end of words that correspond to the main concept of the target image. This suggests that such models are able to isolate the target word forms from fluent speech and thus segment their inputs into sub-units. In the following experiment we test if a RNN-based VGS network can reliably map isolated word-like units to their visual referents and explore the factors that could influence such mapping.

\subsection{Isolated Word Mapping}
\label{doc:word-recognition_isolated-word-mapping}

We selected a set of 80 words corresponding to the name of 80 object categories in the MSCOCO data set.\footnote{List available at \url{https://github.com/amikelive/coco-labels/blob/master/coco-labels-2014_2017.txt}}
We expect our model to be very efficient with the selected 80 words, as these are the main objects featured in MSCOCO. We generated speech signals for these 80 isolated words using Google's TTS system and then extracted MFCC features for each of the generated words. We evaluate the ability of the model to rank images containing an object instance corresponding to the target word among the first 10 images (P@10).\footnote{Evaluation is performed on the test set containing 5000 images} Contrary to \cite{Chrupala2017} who uses Recall@k, we use Precision@k as there are several images that correspond to a single target word. It is to be noted that at training time, the network was only given full captions and not isolated words. Thus, if the network is able to retrieve images featuring instances of the target word, it shows that implicit segmentation was carried out at training time. 

\begin{figure}[htbp!]
	\centering
	\includegraphics[width=0.40\textwidth]{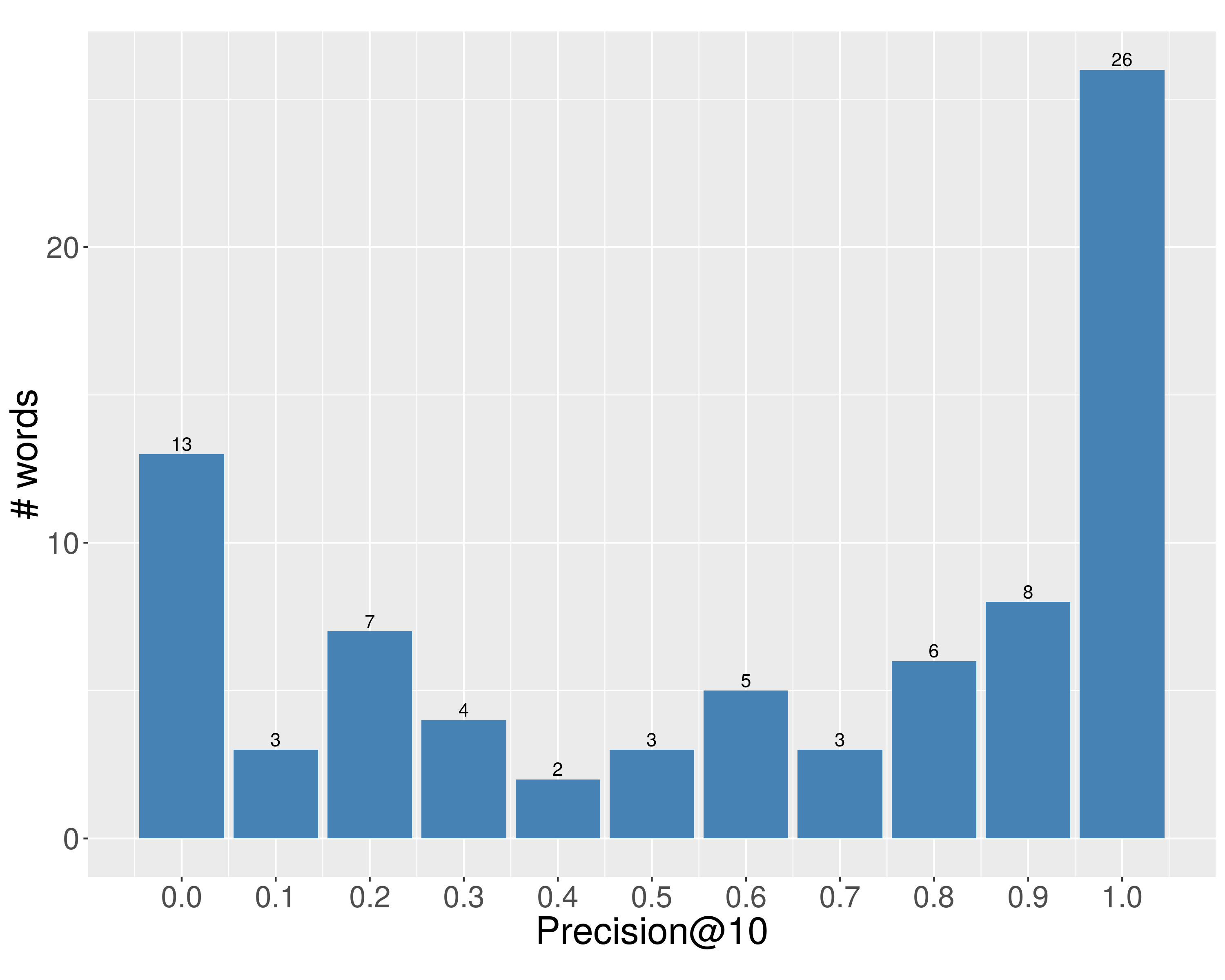}
	\caption{Precision@10 for the 80 isolated words corresponding to MSCOCO categories.}
	\label{fig:isolated_word-recall}
\end{figure}

Results are shown in Figure \ref{fig:isolated_word-recall}. 40 words out of the 80 target words have a P@10$\geq0.8$. This shows that the network is able to map isolated words to their visual referent despite never having seen them in isolation and that the network implicitly segmented its input into sub-units.

\subsection{Factors Influencing Word Mapping}
\label{doc:word-recognition_factors}
We explore here the factors that could come at play in the recognition of isolated words. We explore $2$ types of factors: speech related factors and image related factors. For the former we consider word frequency (Word Freq.) and word length (\# syllables). Concerning image related factors we consider object instances frequency in the images (Avg. Freq.), average number of neighbouring object instances (Avg. Neighbour), average area of each object (Avg. Size). Results are shown in Table \ref{tab:tab_correlation}. We observe a weak negative correlation between precison and average number of neighbouring objects, thus suggesting that objects that have a low number of neighbouring objects are better recognised by the network. It also seems that bigger objects yield better precison than smaller objects as we observe a weak positive correlation. Word frequency seems to play an important role as we observe a moderate positive correlation. However, we observe no correlation between precison and the length of the target words nor with object frequency in the images. Correlation values, however, remain relatively low, suggesting some other factors could also influence word recognition.

\begin{table}[]
    \resizebox{\linewidth}{!}{
    \begin{tabular}{|c|c|c|c|c|}
    \hline
    \multicolumn{2}{|c|}{} & \textbf{Spearman's $\rho$} & \textbf{p-value} & \textbf{Effect} \\ \hline\hline
    \multirow{3}{*}{\textbf{Images}} & Avg. Neighbour & -0.3906 & 0.0003 *** & Weak \\ \cline{2-5} 
     & Avg. Size & 0.3154 & 0.0043 ** & Weak \\ \cline{2-5} 
     & Avg. Freq & 0.1187 & 0.4675 & None \\ \hline\hline
    \multirow{2}{*}{\textbf{Text}} & Word Freq. & 0.5514 & 1.148e-07 *** & Moderate \\ \cline{2-5} 
     & \# Syllables & -0.1211 & 0.2844 & None \\ \hline
    \end{tabular}
    }
    \caption{Factors influencing word recognition performance in our model. Spearman's $\rho$ between Precision@10 and mentioned variables as well as p-value.}
    \label{tab:tab_correlation}
\end{table}

\begin{figure*}[htbp!]
	\begin{minipage}{.50\linewidth}
		\centering
		\subfloat[]{\label{fig:ablation_global-smooth}\includegraphics[width=1\textwidth]{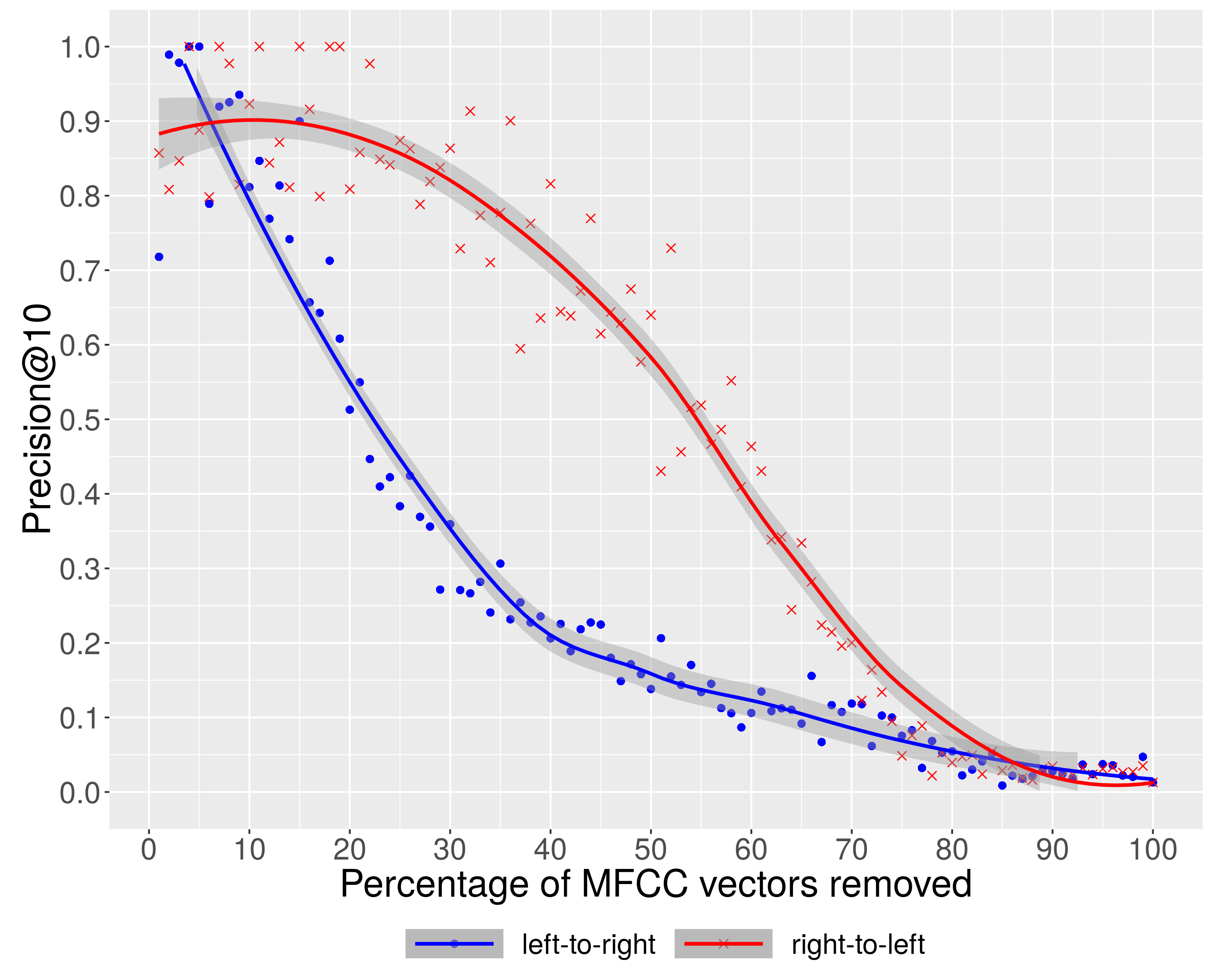}}
	\end{minipage}
    \begin{minipage}{.50\linewidth}
		\centering
		\subfloat[]{\label{fig:ablation_giraffe}\includegraphics[width=1\textwidth]{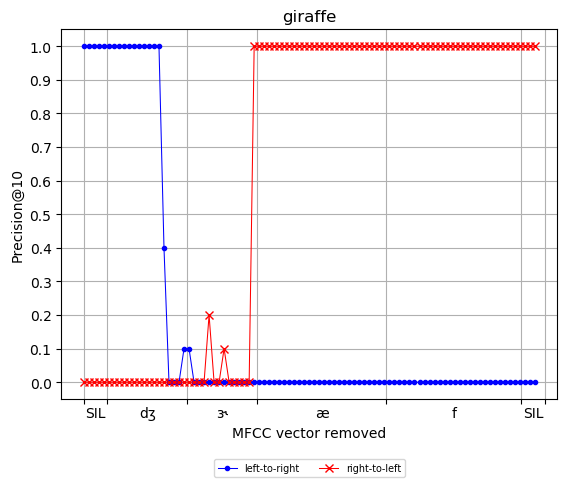}}
	\end{minipage}
	\caption{\ref{fig:ablation_global-smooth} Evolution of Precision@10 averaged over 80 test words as a function of the percentage of MFCC vector removed for each word. \ref{fig:ablation_giraffe} Evolution of Precision@10 for each ablation step of the word ``giraffe", with time-aligned phonemic transcription /\textipa{\textdyoghlig\textrhookrevepsilon\ae f}/ at the bottom. ``\textsc{SIL}" signals silences. For both \ref{fig:ablation_global-smooth} and \ref{fig:ablation_giraffe}, blue line displays scores when ablation was carried out left-to-right, meaning that at any given part on the blue curve, model has only had access to the rightmost part of the word. (e.g. /\textipa{\textrhookrevepsilon  \ae f}/ without initial /\textipa{\textdyoghlig}/). Red line displays scores when ablation was carried out right-to-left, meaning that at any given part on the red curve, model has only had access to the leftmost part of the word. (e.g. /\textipa{\textdyoghlig\textrhookrevepsilon}/ without final /\textipa{\ae f}/ ).}
	\label{fig:ablation}
\end{figure*}

\section{Word Activation}
\label{doc:word-activation_intro}
In this section we describe how individual words are activated by the network. 
To do so, we perform an ablation experiment (similar to that of \citet{Grosjean1980} which was conducted on humans) where the neural model is inputted only with a truncated version of the 80 target words (see Section 
\ref{doc:word-activation_truncation}).
Such a method is also called \textit{gating} in the literature.

\subsection{Gating}
\label{doc:word-activation_truncation}
The gating paradigm \textit{``involves the repeated presentation of a spoken stimulus (in this case, a word) such that its duration from onset is increased with each successive presentation"} \cite{Cotton1984}. In our case, it means the neural model is fed with truncated version of a target word, each truncated version comprising a larger part of the target word. Truncation is either done left-to-right (model only has access to the end of the word) or right-to-left (model only has access to the beginning of the word). Truncation is operated on the MFCC vectors computed for each individual word, meaning that MFCC vectors are iteratively removed either from the beginning of the word or the end of the word, but not from both sides at the same time. Each truncated version of the word is then fed to the speech encoder which outputs an embedding vector. As in our previous experiment, model is evaluated in terms of P@10. 

\textsc{cohort} model, in its initial version \cite{marslenwillow87_cohort}, stipulates that word onset plays a crucial role in word recognition whereas other models of spoken word recognition give less importance to word onset. This importance of exact word onset matching was later revised in later \textsc{cohort} models. The aim of this experiment is to test whether word onset plays a role in word recognition for the network or not. If it is the case, we expect the network to fail recovering images of the target word if the word is truncated left-to-right.

Figure \ref{fig:ablation_global-smooth} shows evolution of P@10 averaged over the 80 test words. As can be seen from the graph, precision evolves differently according to which part of the word was truncated. When the target words are truncated left-to-right, precision drops quickly. However, when truncated right-to-left, precision remains high before gradually dropping. These results show that the model is robust to truncation when it is carried out right-to-left but not when it is carried out left-to-right. Figure \ref{fig:ablation_giraffe} shows the evolution of P@10 for one of the target words (``giraffe"). When MFCC vectors corresponding to the first phoneme are removed (/\textipa{\textdyoghlig}/), precision plummets from 1 to 0. However, when MFCC vectors belonging to the end of the word are removed, precision plateaus at 1 until /\textipa{\textrhookrevepsilon}/ is reached and then plunges to 0. This shows the model successfully retrieved giraffe images when only prompted with /\textipa{\textdyoghlig\textrhookrevepsilon}/ but not when prompted with /\textipa{\textrhookrevepsilon\ae f}/ even though the latter comprises a longer part of the target word.

These results suggest that the model does not rely on a vague acoustic pattern to activate the semantic representation of a given concept, but needs to have access to the first phoneme in order to yield an appropriate representation.

\subsection{Activated Pseudo-Words}
\label{doc:word-activation_pseudo-words}

\begin{figure}[h]
	\centering
	\includegraphics[width=0.48\textwidth]{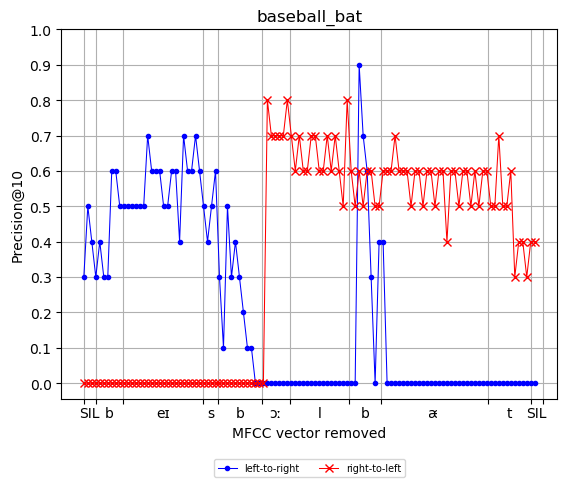}
	\caption{Evolution of P@10 for each ablation step of the word ``baseball bat" with time aligned phonemic transcription /\textipa{beizb\textopeno:l}\#\textipa{b\ae t}/ at the bottom.
	}
	\label{fig:rank_baseballbat}
\end{figure}

Such ablation experiments also enables us to infer on what units the network relies to make its predictions. Indeed, Figure \ref{fig:rank_baseballbat} allows us to see what are the pseudo-words that were internalised by the network for the word ``baseball bat". When truncation is done left-to-right (blue curve), we notice that at the beginning precision is quite high ($\approx0.6$), then reaches $0$ when only /\textipa{\textopeno:lb\ae t}/ is left, but suddenly increases up to $0.9$ when the only part left is /\textipa{b\ae t}/. This suggests that the network mapped both ``baseball bat" as a whole and ``bat" as referring to the same object. We observed the same pattern for the word ``fire hydrant" where both ``fire hydrant" and ``hydrant" are mapped to the same object. 

However, Figure \ref{fig:ablation_giraffe} shows that when only prompted with /\textipa{\textdyoghlig\textrhookrevepsilon}/ the network manages to find pictures of giraffes. This suggests that the pseudo-words internalised by the network could be /\textipa{\textdyoghlig\textrhookrevepsilon\ae f}/ as a whole but might also be /\textipa{\textdyoghlig\textrhookrevepsilon}/. We thus need to take caution when stating that the network has isolated words, as the words internalised by the network might not always match the human gold reference.

\subsection{Gradual or Abrupt Activation?}
\label{doc:word-activation_linear-activation}
\begin{figure*}[h]
	\begin{minipage}{.33\linewidth}
		\centering
		\subfloat[]{\label{fig:cos_giraffe0}\includegraphics[width=1\textwidth]{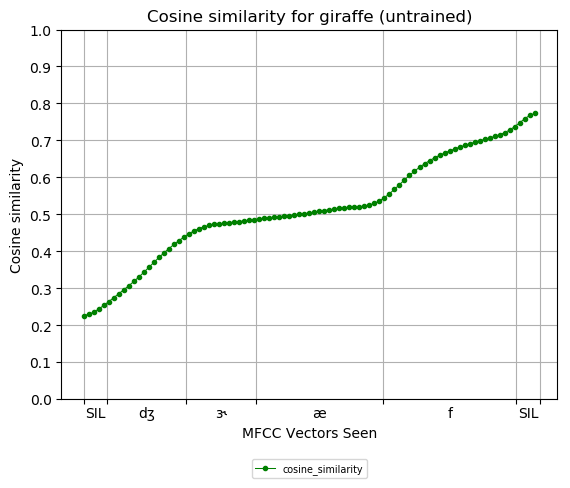}}
	\end{minipage}
	\hspace{-0.2cm}
	\begin{minipage}{.34\linewidth}
		\centering
		\subfloat[]{\label{fig:cos_peak_giraffe}\includegraphics[width=1\textwidth]{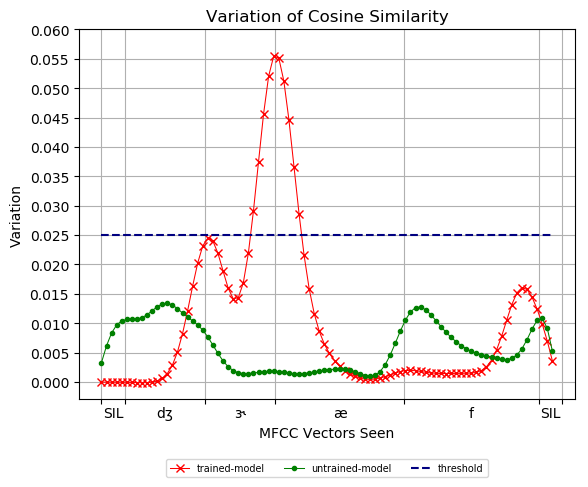}}
	\end{minipage}
	\hspace{-0.2cm}
	\begin{minipage}{.33\linewidth}
		\centering
		\subfloat[]{\label{fig:cos_giraffe12}\includegraphics[width=1\textwidth]{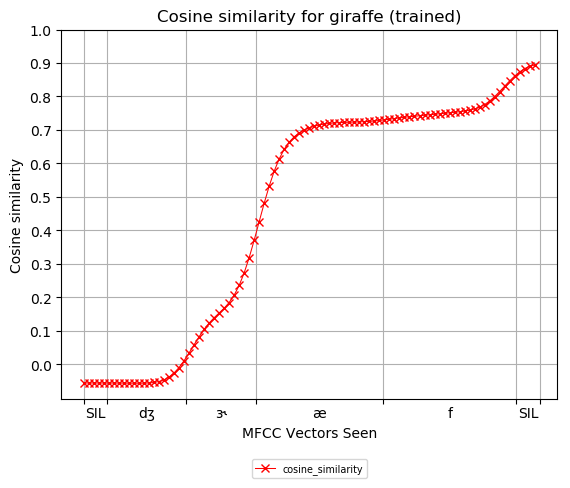}}
	\end{minipage}
	\caption{Figure \ref{fig:cos_giraffe0} shows evolution of the cosine similarity between the embeddings produced for each truncated version of the target word and the embedding for the full word using a model with randomly initialised weights. Figure \ref{fig:cos_giraffe12} shows the same measure with the embeddings produced by a trained model. Figure \ref{fig:cos_peak_giraffe} shows peaks indicating the inflection points of curve \ref{fig:cos_giraffe0} (green) and  \ref{fig:cos_giraffe12} (red). For our experiments, we only considered inflection point to be significant if the resulting peak was higher than 0.025 (blue).}
	\label{fig:cos_sim}
\end{figure*}
Figure \ref{fig:ablation_giraffe} shows that removing or adding one MFCC vector may yield large differences in the network performance. Precision decreases steeply and not steadily. This suggests that little acoustic differences yield wide differences in the final representation. Thus, in this section we analyse how representation is being constructed over time and explore if some MFCC vectors play a more important role than others in the activation of the final representation.

We progressively let the network see more and more of the MFCC vectors composing the word, iteratively feeding it with MFCC vectors starting from the beginning of the word until the network has had access to the full word. We then compute the cosine similarity between the embedding computed for each of the truncated version of the word and the embedding corresponding to the full word. 
The closer the cosine similarity is to $1$, the more similar the two representations are. Thus, if each MFCC vector equally contributes to the final representation of the word, we expect cosine similarity to evolve linearly. However, if some MFCC vectors have a determining factor in the final representation we expect cosine similarity to evolve in steps rather than linearly. To detect steps that could occur in the evolution of cosine similarity, we approximate its derivative by computing first order difference. High steps should thus translate into peaks (e.g. Figure \ref{fig:cos_peak_giraffe}). We compute the evolution of cosine similarity for the 80 target words encoded with the best trained model (e.g. Figure \ref{fig:cos_giraffe12}) and also consider a baseline evolution by encoding the 80 target words with an untrained model (e.g. Figure \ref{fig:cos_giraffe0}).\footnote{Thus consisting only of randomly initialised weights} To avoid micro-steps of yielding peaks and thus creating noise, we smooth cosine evolution curves with a gaussian filter. We consider peaks higher than 0.025 as translating a high step in the evolution of cosine similarity.

On average, they are 1.35 peaks per word for the trained model against 0.1 peak per word for our baseline condition (untrained model), showing that cosine evolution is linear in the latter but not in the former. Thus, in our baseline condition (untrained model), each MFCC vector equally contributes to the final representation, whereas in our trained model some MFCC vectors are more decisive for the final representation than others. Indeed, some MFCC vectors trigger a high step in the cosine evolution suggesting that the embedding suddenly gets closer to its final value.  Figure \ref{fig:cos_giraffe12} shows the evolution of the cosine similarity for the word ``giraffe''. As it can be seen, cosine similarity does not tend linearly towards $1$, but rather evolves in steps. Adding the MFCC vectors corresponding to the transition from /\textipa{\textrhookrevepsilon}/ to  /\textipa{\ae}/ triggers a large difference in the embedding as the cosine similarity suddenly jumps to a higher value, showing it is getting closer to its final representation. However, cosine similarity plateaus once /\textipa{\ae}/ is reached up until final silence, suggesting the final  /\textipa{\ae f}/ plays little to no role in the final representation of the word.

\section{Word Competition}
\label{doc:word-competition_intro}

\begin{figure*}[h]
    \begin{minipage}{.50\linewidth}
		\centering
		\subfloat[]{\label{fig:competition_meat-meter}\includegraphics[width=1\textwidth]{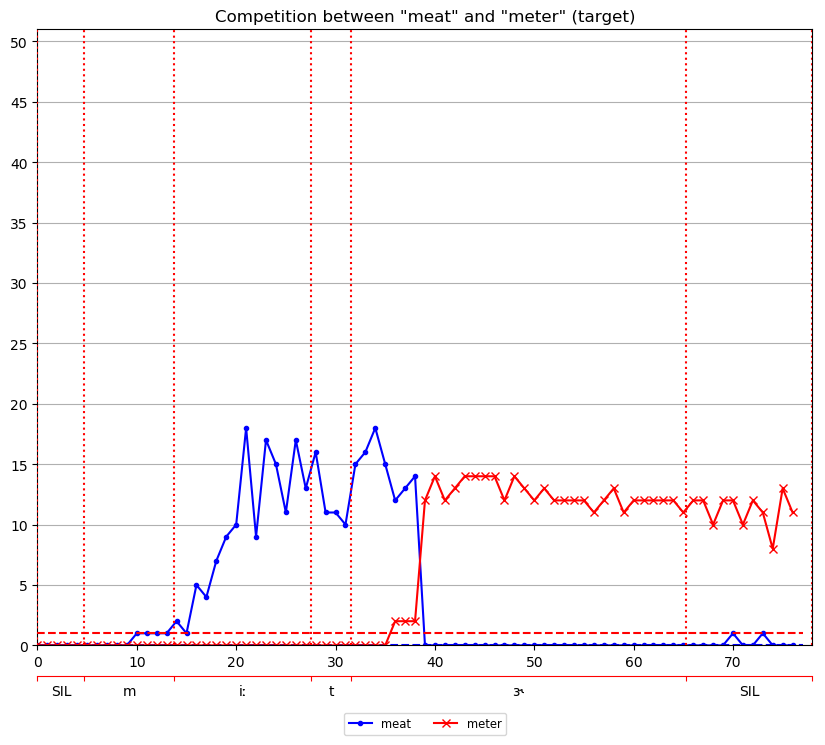}}
	\end{minipage}
	\begin{minipage}{.50\linewidth}
		\centering
		\subfloat[]{\label{fig:competition_player-plate}\includegraphics[width=1\textwidth]{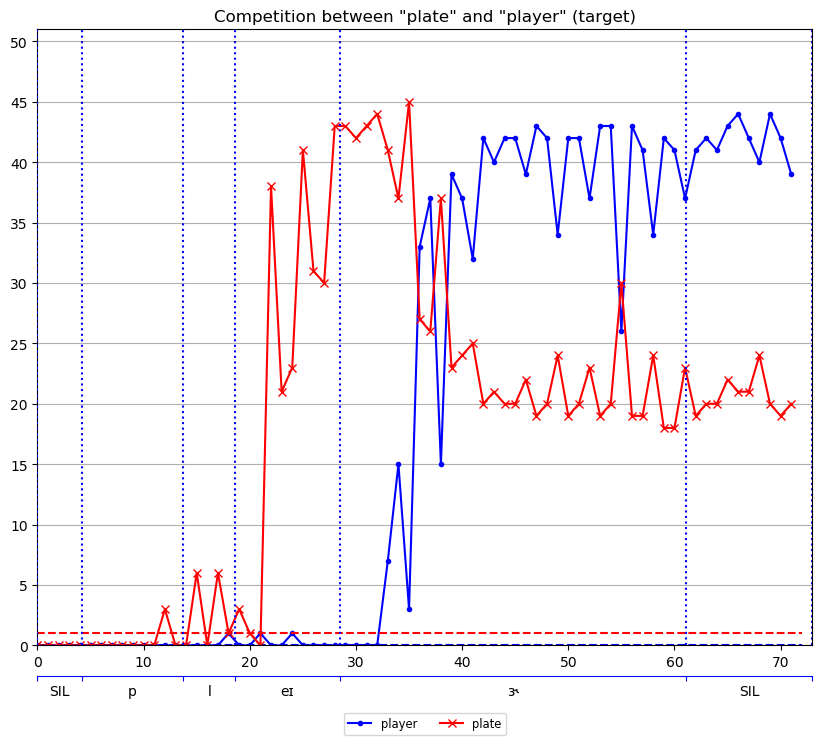}}
	\end{minipage}
	\caption{Illustration of lexical competition between \ref{fig:competition_meat-meter} ``meat" and ``meter" (target) and \ref{fig:competition_player-plate} ``plate" and ``player" (target). Numbers in 1$^{st}$ x-axis corresponds to the number of MFCC frames; 2$^{nd}$ x-axis corresponds to time-aligned phonemic transcription of the target word; y-axis shows number of images for which at least one caption (out of 5) contains the target or competitor word. Vertical colour bars are projection of phoneme boundaries of the target word. Horizontal colour bars show chance score for each word (\textless 2).
	}
	\label{fig:competition}
\end{figure*}
As presented in Section \ref{doc:related-work_human-activation}, some linguistic models assume that the first phoneme of target word activates all the words starting by the same phoneme. The words that are activated but which do not correspond to the target words are called ``competitors". As the listener perceives more and more of the target word some competitors are deactivated as they do not match what is being perceived. For example, considering the following lexicon: /\textipa{be\textsci bi}/ (baby), /\textipa{be\textsci z\textsci k}/ (basic), and /\textipa{be\textsci zb\textopeno:l}/ (baseball), the first sound /\textipa{b}/ would activate all three words, once /\textipa{be\textsci z}/ is reached, ``baby" would not be considered a competitor anymore, and once /be\textsci zb/ is reached the only word activated would be ``baseball" as it is the only word whose beginning corresponds the perceived sounds.

We test if the network displays such lexical competition patterns. To do so, we select a set of 29 word pairs according to the following criteria: i) words should at least appear 400 times or more in the captions of the training set, so that the network would have been able to learn a mapping between this word and its referent; ii) words forming a pair should at least start with the same phoneme;\footnote{Phonemic transcription found in CMU Pronouncing Dictionnary was used} iii) words should not be synonyms and clearly refer to a different visual object (thus excluding pairs such as ``motorcycle" and ``motorbike"). 

For each word pair, we select the longest word as target and  progressively let the network see more and more of the MFCC vectors composing this word (as in Section \ref{doc:word-activation_linear-activation}). At each time step the network produces an embedding, which we use to rank the images from the closest matching image to the least matching image.\footnote{That is, we compute the cosine distance between the embedding produced at time step $t$ and all the images (5000) of our collection.} Then, for the 50 closest matching images, we check if at least one of the caption contains either the target word or the competitor. As the competitor is embedded in the longer word, we expect the network to produce an embedding close to that of the competitor at the beginning and then when the acoustic signal does not match the competitor anymore, we expect the network to be able to find only the target word.

Figure \ref{fig:competition} shows example of competition between two word pairs. Figure \ref{fig:competition_meat-meter} shows that when prompted with the beginning of the word ``meter" /\textipa{mi:t}/ the representation activated by the network is close to that of ``meat" as the closest maching images's captions contain the word ``meat". Representation of the word ``meter" seems to be activated only when /\textipa{\textrhookrevepsilon}/ is reached, and consequently triggers the total deactivation of the word ``meat". Figure \ref{fig:competition_player-plate} displays a different pattern. As in the previous example, the beginning of the word ``player" /\textipa{ple\textsci}/ triggers the activation of the word ``plate". When /\textipa{\textrhookrevepsilon}/ is reached, the target word becomes activated and competitor ``plate" starts to deactivate. However, the deactivation is not full, so that when the whole word ``player" is entirely processed by the network, the word ``plate" still remains highly activated.
\textsc{(revised) cohort} \cite{marslenwillow78_cohort, marslenwillow87_cohort} and \textsc{trace} \cite{mcclelland_trace} both state that competing words are all activated at the same time, that is when the first phoneme is perceived. However here, the two competing words are activated sequentially but not at the same time. Also, in some cases, competing words that do not match the input anymore still remain highly activated. 

\section{Conclusion}
\label{concl}

In this paper, we analysed the behaviour of a model of VGS and showed that a RNN-based model of VGS is able to map isolated words to their visual referents. This result is in line with previous results, such as that of \citet{Harwarth19_subword} which uses a CNN-based network. This shows that such models perform an implicit segmentation of the spoken input in order to extract the target words. However, the mechanism by which implicit segmentation is carried out and what cues are being used is still to be explained. We also demonstrated that not all words are equally well recognised and  showed that word frequency and number of neighbouring object in an image partly explain this phenomenon.

Also, we introduced a methodology originating from linguistics to analyse the representation learned by neural networks: the gating paradigm. This enabled us to show that the beginning of a word can activate the representation of a given concept (e.g. /\textipa{\textdyoghlig\textrhookrevepsilon}/ for ``giraffe"). We explain this by the fact that the network has to handle a very small lexicon, where word forms rarely overlap and thus the network needs not see the full word to make its decision. More importantly, we showed that the network needs to have access to the first phoneme in order to activate the representation of the target word, thus showing that it does not respond to a vague acoustic pattern. Word onsets thus play a crucial role in the process of word activation and recognition for our network. Though word onsets are also important for humans, they are not as crucial as for our network. Indeed, humans are able to recover the missing information. In future work, we would like to test if sentential context has an effect in word recognition. We also demonstrated that our model is able to map multiple pseudo-words to the same referent such as humans do (Section \ref{doc:word-activation_pseudo-words}). However, it is not clear how and when acoustics interface with meaning and this still remains an open question.

Finally, we showed that there is a form of lexical competition in the network. Indeed, small words embedded in longer words are activated. However, we showed that, contrary to humans where words sharing the same beginning are all activated at the same time, words are activated sequentially by the network. Also, some stay partially activated eventhough the input does not match that of the activated word.

Ultimately, we would like to highlight the fact that the gating paradigm could also by applied to understand the temporal dynamics of the representations learned by other speech architectures such as those used in speech recognition for instance.

\section*{Acknowledgments}
This work was supported by grants from NeuroCoG IDEX UGA as part of the ``Investissements d'avenir" program (ANR-15-IDEX-02).

%
%

\bibliography{main}
\bibliographystyle{acl_natbib}

%
%

%
%

\newpage
\appendix
\section{Supplementary Material}
\label{sec:supplementary}
\subsection{Precision@10 for the 80 test words}
\label{sec:supplementary-recall}
    \begin{figure}[h!]
    	\begin{minipage}{2\linewidth}
    		\centering
    		\includegraphics[width=1\linewidth]{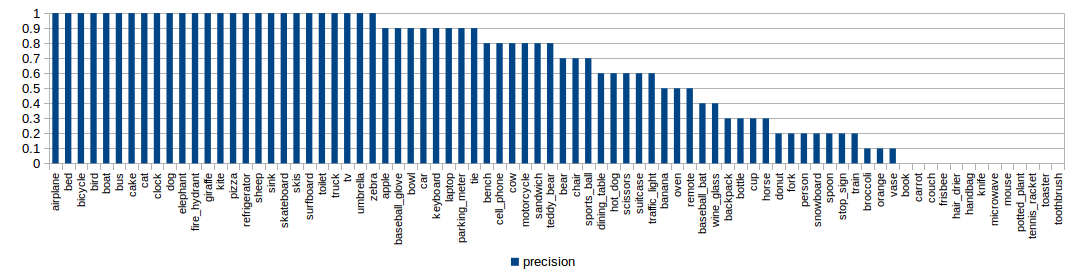}
            \caption*{Figure: Precision@10 for each of the 80 test words}
            \label{fig:sup-mat-recall}
    	\end{minipage}
    \end{figure}
    \begin{figure*}[h!]
\subsection{Gating Samples} 
\label{sec:supplementary-gating}
    \begin{minipage}{.50\linewidth}
    		\centering
    		\subfloat[]{\label{fig:sup-mat-ablation-fork}\includegraphics[width=1\textwidth]{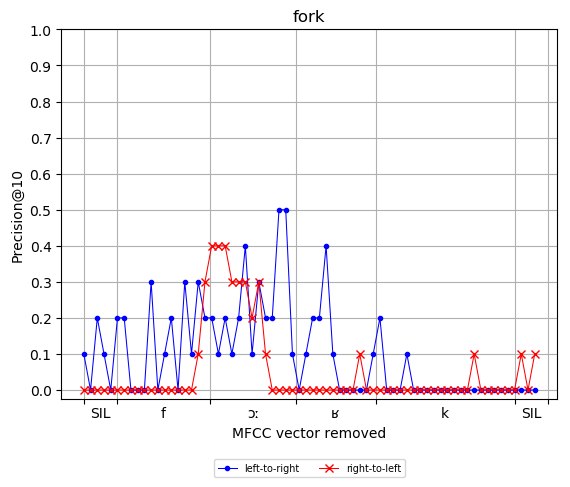}}
    	\end{minipage}
    	\begin{minipage}{.50\linewidth}
    		\centering
    		\subfloat[]{\label{fig:sup-mat-ablation-microwave}\includegraphics[width=1\textwidth]{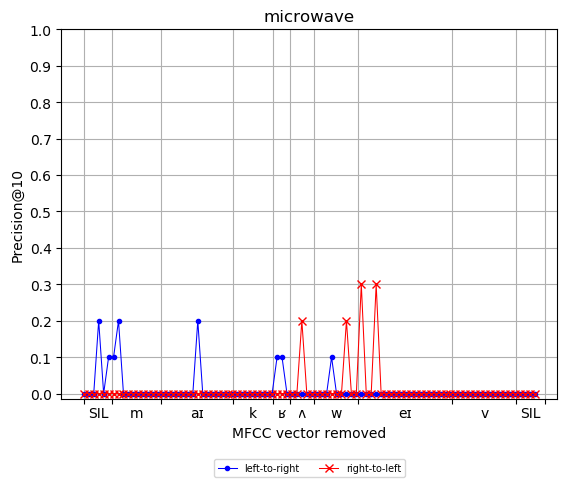}}
    	\end{minipage}
    \end{figure*}
    \begin{figure*}[h]\ContinuedFloat
        \begin{minipage}{.50\linewidth}
    		\centering
    		\subfloat[]{\label{fig:sup-mat-ablation-motorcycle}\includegraphics[width=1\textwidth]{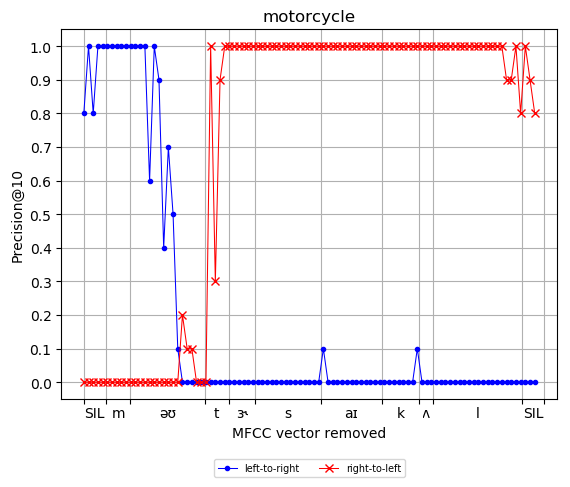}}
    	\end{minipage}
    	\begin{minipage}{.50\linewidth}
    		\centering
    		\subfloat[]{\label{fig:sup-mat-ablation-pizza}\includegraphics[width=1\textwidth]{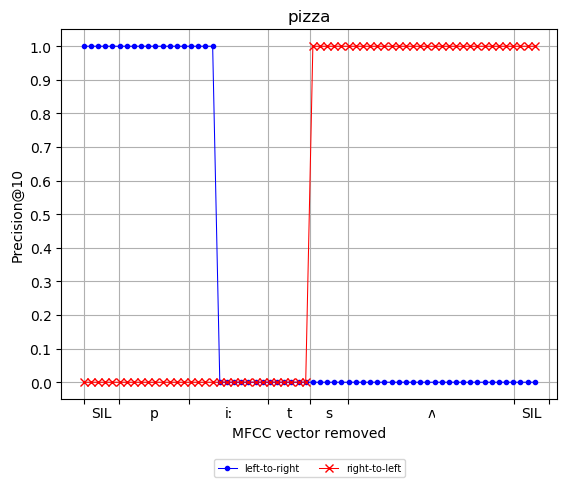}}
    	\end{minipage}
    \end{figure*}
    \begin{figure*}[h]\ContinuedFloat
        \begin{minipage}{.50\linewidth}
    		\centering
    		\subfloat[]{\label{fig:sup-mat-ablation-scissors}\includegraphics[width=1\textwidth]{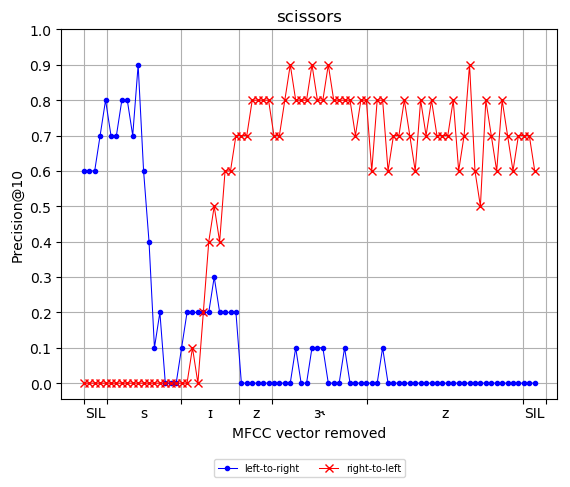}}
    	\end{minipage}
    	\begin{minipage}{.50\linewidth}
    		\centering
    		\subfloat[]{\label{fig:sup-mat-ablation-sink}\includegraphics[width=1\textwidth]{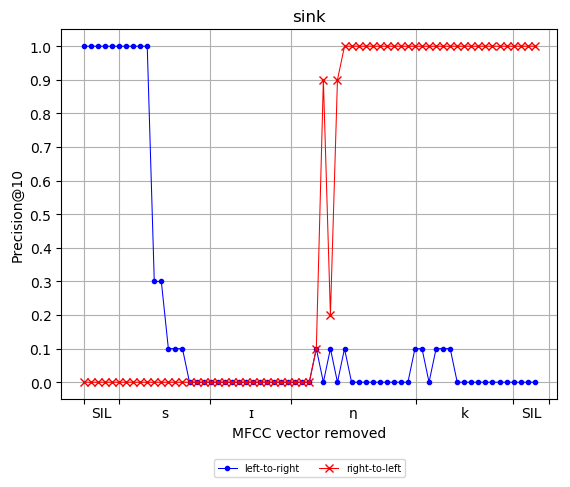}}
    	\end{minipage}
    \end{figure*}
    \begin{figure*}[h]\ContinuedFloat
        \begin{minipage}{.50\linewidth}
    		\centering
    		\subfloat[]{\label{fig:sup-mat-ablation-surfboard}\includegraphics[width=1\textwidth]{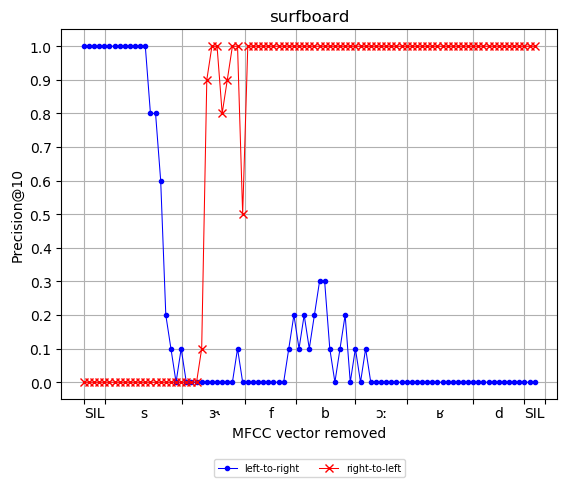}}
    	\end{minipage}
    	\begin{minipage}{.50\linewidth}
    		\centering
    		\subfloat[]{\label{fig:sup-mat-ablation-toilet}\includegraphics[width=1\textwidth]{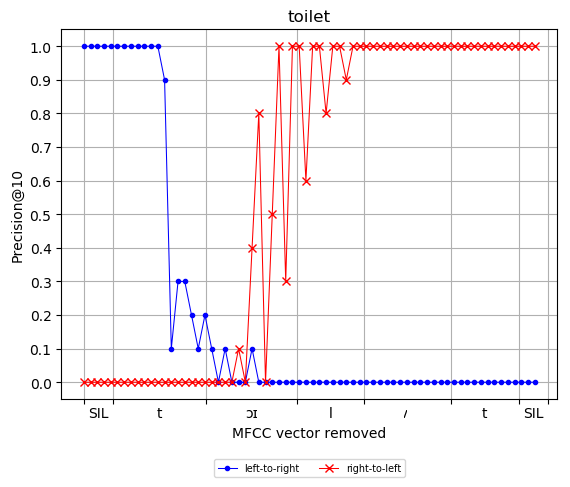}}
    	\end{minipage}
    \end{figure*}
    \begin{figure*}[h]\ContinuedFloat
        \begin{minipage}{.50\linewidth}
    		\centering
    		\subfloat[]{\label{fig:sup-mat-ablation-traffic_light}\includegraphics[width=1\textwidth]{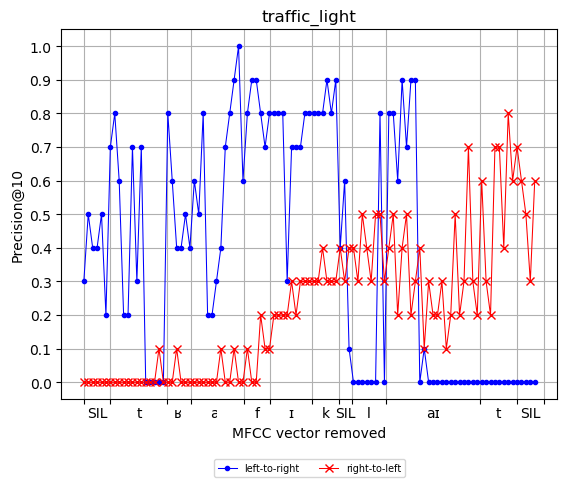}}
    	\end{minipage}
    	\begin{minipage}{.50\linewidth}
    		\centering
    		\subfloat[]{\label{fig:sup-mat-ablation-train}\includegraphics[width=1\textwidth]{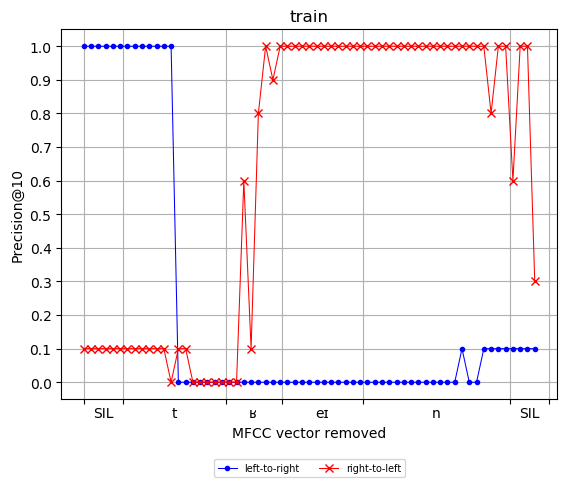}}
    	\end{minipage}
    \end{figure*}
    \begin{figure*}[h]\ContinuedFloat
        \begin{minipage}{.50\linewidth}
    		\centering
    		\subfloat[]{\label{fig:sup-mat-ablation-wine_glass}\includegraphics[width=1\textwidth]{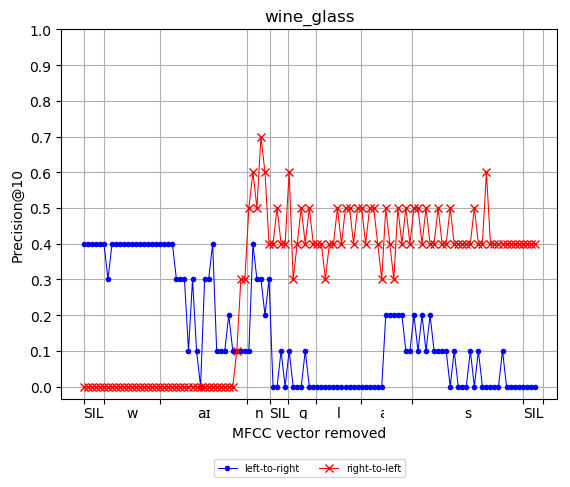}}
    	\end{minipage}
    	\begin{minipage}{.50\linewidth}
    		\centering
    		\subfloat[]{\label{fig:sup-mat-ablation-zebra}\includegraphics[width=1\textwidth]{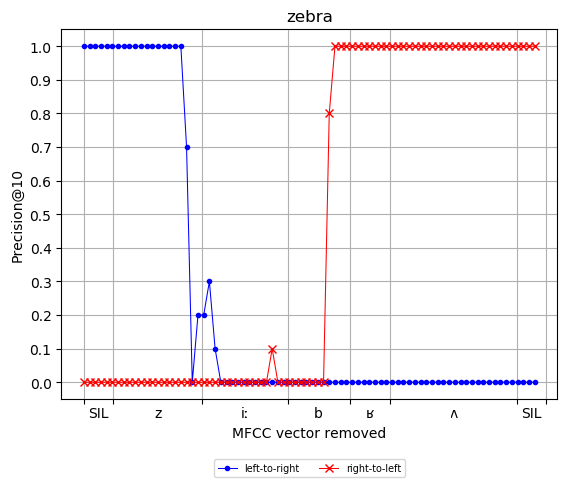}}
    	\end{minipage}
		\caption*{Figure: Evolution of Precision@10 for each ablation step for the words (a) ``fork", (b) ``microwave", (c) ``motorcycle", (d) ``pizza", (e) ``scissors", (f) ``sink", (g) ``surfboard, (h) ``toilet", (i) ``traffic light", (j) ``train", (k) ``wine glass", (l) ``zebra". It should be noted that ``fork" and ``microwave" do not display the same behaviour as the other words, this could be explained by the fact these two words both have a very low Precision@10.}
    \end{figure*}
        
\begin{figure*}[h]
\subsection{Activation Sample}
\label{sec:supplementary-cosine}
	\begin{minipage}{.33\linewidth}
		\centering
		\subfloat[]{\label{fig:sup-mat-cosine-untrained-fork}\includegraphics[width=1\textwidth]{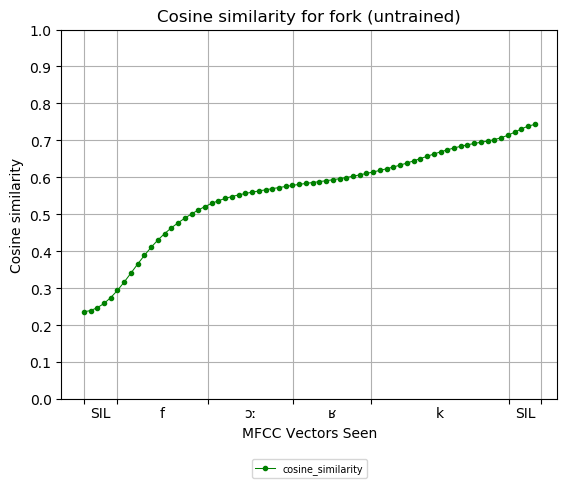}}
	\end{minipage}
	\hspace{-0.2cm}
	\begin{minipage}{.34\linewidth}
		\centering
		\subfloat[]{\label{fig:sup-mat-cosine-diff-fork}\includegraphics[width=1\textwidth]{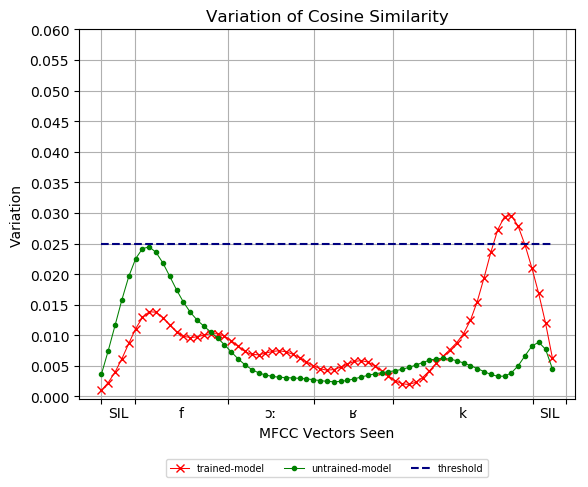}}
	\end{minipage}
	\hspace{-0.2cm}
	\begin{minipage}{.33\linewidth}
		\centering
		\subfloat[]{\label{fig:sup-mat-cosine-trained-fork}\includegraphics[width=1\textwidth]{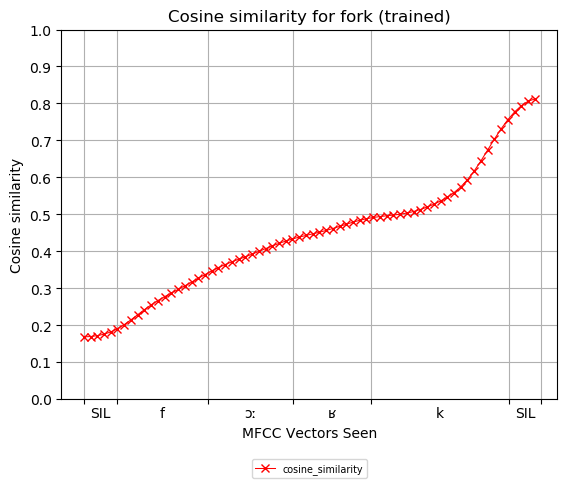}}
	\end{minipage}
	\caption{Evolution of cosine similarity for the word ``fork". Figure \ref{fig:sup-mat-cosine-diff-fork} shows peaks indicating the inflection points of curve \ref{fig:sup-mat-cosine-untrained-fork} (untrained model, green) and \ref{fig:sup-mat-cosine-trained-fork} (trained model, red)}
	\label{fig:sup-mat-cosine-fork}
\end{figure*}
        
\begin{figure*}[h]

	\begin{minipage}{.33\linewidth}
		\centering
		\subfloat[]{\label{fig:sup-mat-cosine-untrained-microwave}\includegraphics[width=1\textwidth]{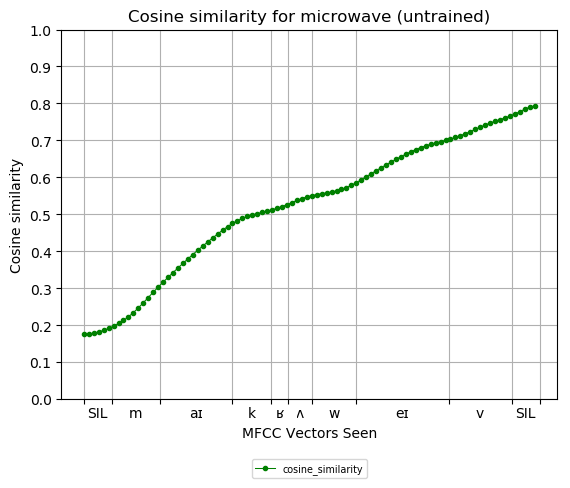}}
	\end{minipage}
	\hspace{-0.2cm}
	\begin{minipage}{.34\linewidth}
		\centering
		\subfloat[]{\label{fig:sup-mat-cosine-diff-microwave}\includegraphics[width=1\textwidth]{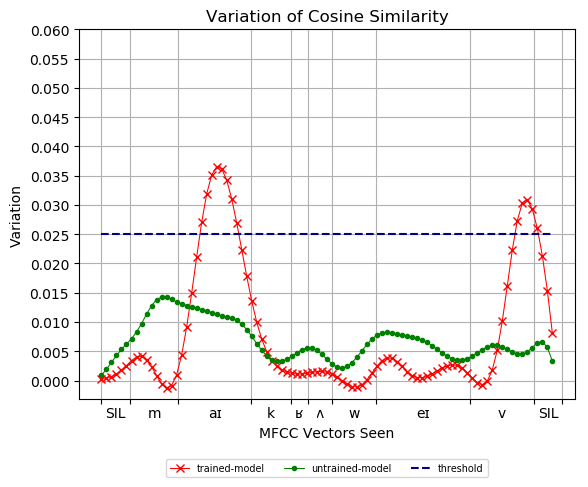}}
	\end{minipage}
	\hspace{-0.2cm}
	\begin{minipage}{.33\linewidth}
		\centering
		\subfloat[]{\label{fig:sup-mat-cosine-trained-microwave}\includegraphics[width=1\textwidth]{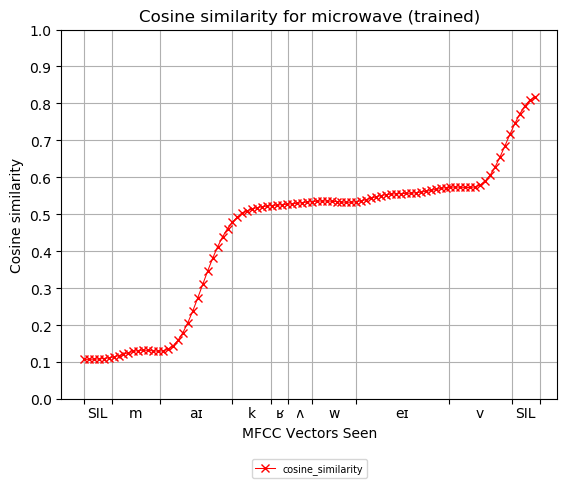}}
	\end{minipage}
	\caption{Evolution of cosine similarity for the word ``microwave". Figure \ref{fig:sup-mat-cosine-diff-microwave} shows peaks indicating the inflection points of curve \ref{fig:sup-mat-cosine-untrained-microwave} (untrained model, green) and \ref{fig:sup-mat-cosine-trained-microwave} (trained model, red)}
	\label{fig:sup-mat-cosine-microwave}
\end{figure*}
        
\begin{figure*}[h]

	\begin{minipage}{.33\linewidth}
		\centering
		\subfloat[]{\label{fig:sup-mat-cosine-untrained-motorcycle}\includegraphics[width=1\textwidth]{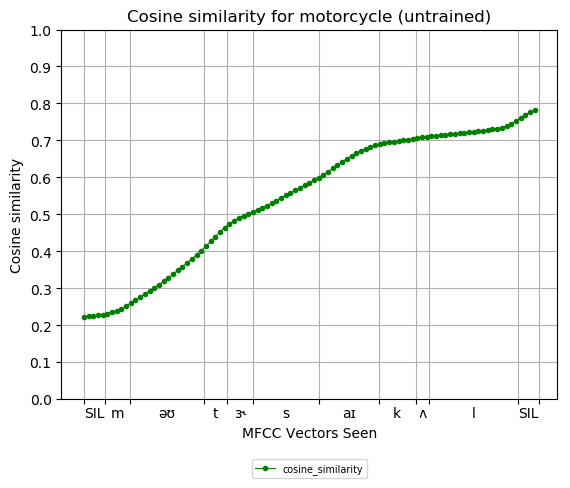}}
	\end{minipage}
	\hspace{-0.2cm}
	\begin{minipage}{.34\linewidth}
		\centering
		\subfloat[]{\label{fig:sup-mat-cosine-diff-motorcycle}\includegraphics[width=1\textwidth]{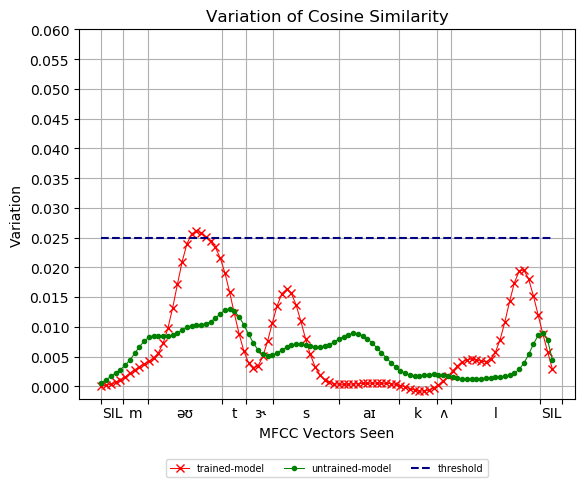}}
	\end{minipage}
	\hspace{-0.2cm}
	\begin{minipage}{.33\linewidth}
		\centering
		\subfloat[]{\label{fig:sup-mat-cosine-trained-motorcycle}\includegraphics[width=1\textwidth]{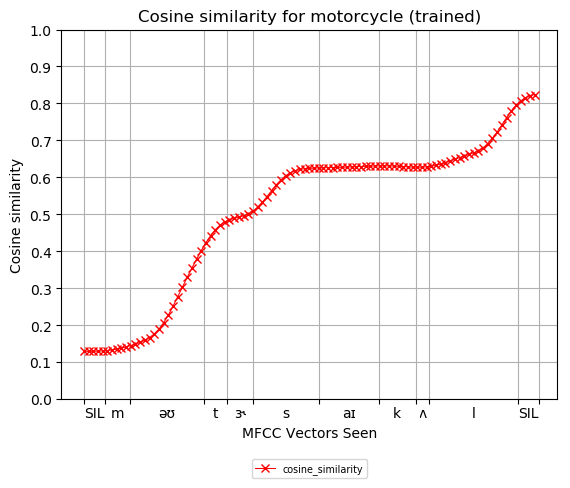}}
	\end{minipage}
	\caption{Evolution of cosine similarity for the word ``motorcycle". Figure \ref{fig:sup-mat-cosine-diff-motorcycle} shows peaks indicating the inflection points of curve \ref{fig:sup-mat-cosine-untrained-motorcycle} (untrained model, green) and \ref{fig:sup-mat-cosine-trained-motorcycle} (trained model, red)}
	\label{fig:sup-mat-cosine-motorcycle}
\end{figure*}
        
\begin{figure*}[h]

	\begin{minipage}{.33\linewidth}
		\centering
		\subfloat[]{\label{fig:sup-mat-cosine-untrained-pizza}\includegraphics[width=1\textwidth]{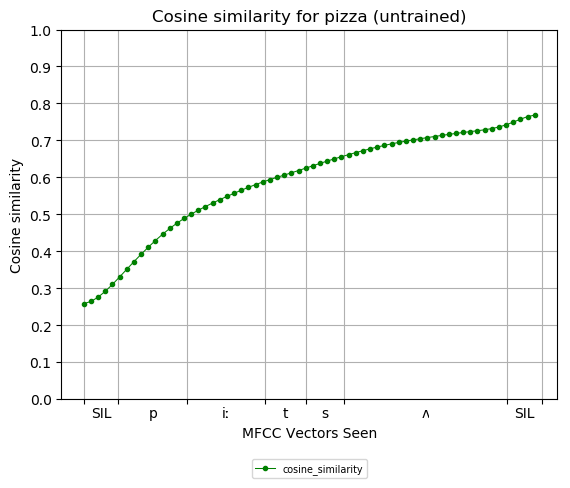}}
	\end{minipage}
	\hspace{-0.2cm}
	\begin{minipage}{.34\linewidth}
		\centering
		\subfloat[]{\label{fig:sup-mat-cosine-diff-pizza}\includegraphics[width=1\textwidth]{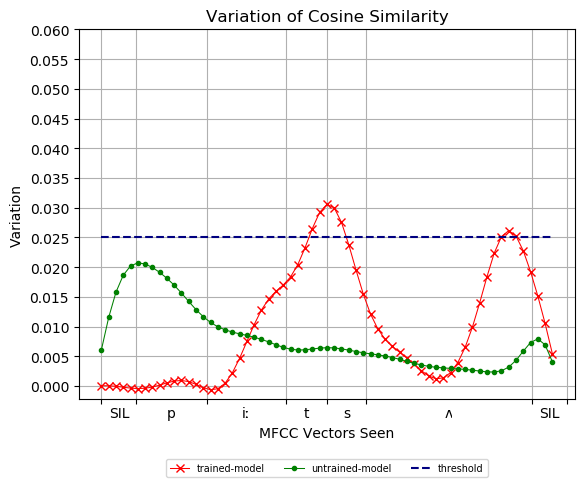}}
	\end{minipage}
	\hspace{-0.2cm}
	\begin{minipage}{.33\linewidth}
		\centering
		\subfloat[]{\label{fig:sup-mat-cosine-trained-pizza}\includegraphics[width=1\textwidth]{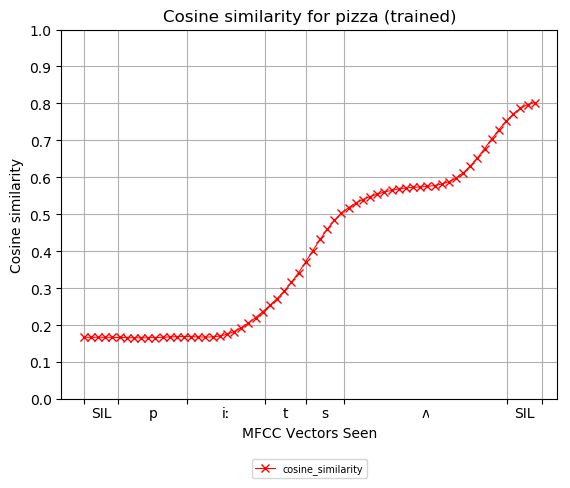}}
	\end{minipage}
	\caption{Evolution of cosine similarity for the word ``pizza". Figure \ref{fig:sup-mat-cosine-diff-pizza} shows peaks indicating the inflection points of curve \ref{fig:sup-mat-cosine-untrained-pizza} (untrained model, green) and \ref{fig:sup-mat-cosine-trained-pizza} (trained model, red)}
	\label{fig:sup-mat-cosine-pizza}
\end{figure*}
        
\begin{figure*}[h]

	\begin{minipage}{.33\linewidth}
		\centering
		\subfloat[]{\label{fig:sup-mat-cosine-untrained-scissors}\includegraphics[width=1\textwidth]{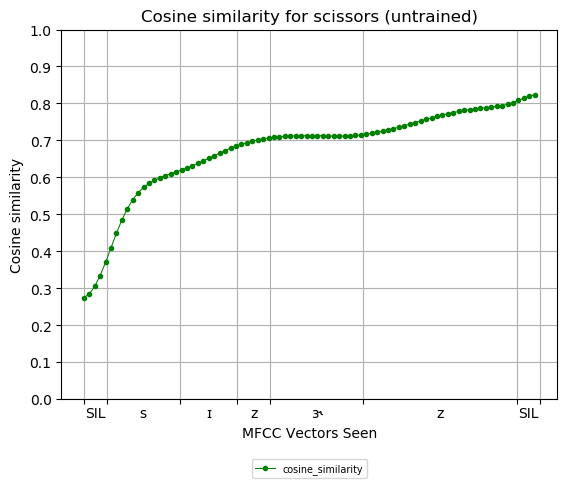}}
	\end{minipage}
	\hspace{-0.2cm}
	\begin{minipage}{.34\linewidth}
		\centering
		\subfloat[]{\label{fig:sup-mat-cosine-diff-scissors}\includegraphics[width=1\textwidth]{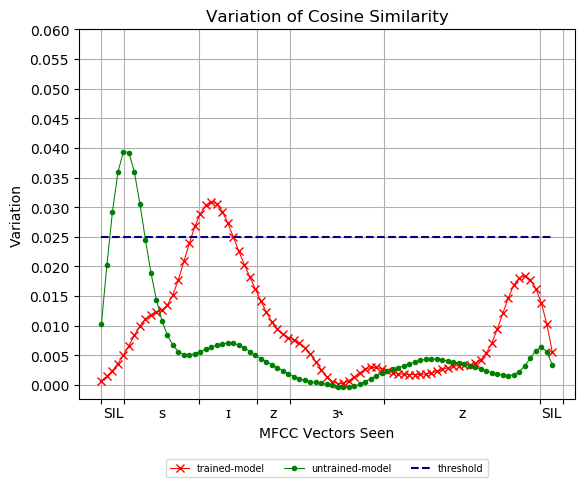}}
	\end{minipage}
	\hspace{-0.2cm}
	\begin{minipage}{.33\linewidth}
		\centering
		\subfloat[]{\label{fig:sup-mat-cosine-trained-scissors}\includegraphics[width=1\textwidth]{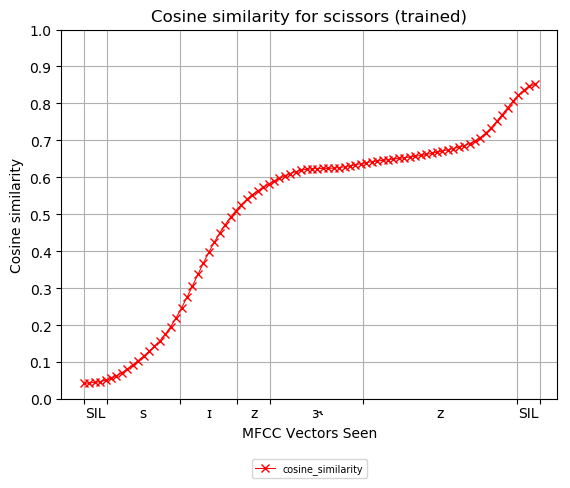}}
	\end{minipage}
	\caption{Evolution of cosine similarity for the word ``scissors". Figure \ref{fig:sup-mat-cosine-diff-scissors} shows peaks indicating the inflection points of curve \ref{fig:sup-mat-cosine-untrained-scissors} (untrained model, green) and \ref{fig:sup-mat-cosine-trained-scissors} (trained model, red)}
	\label{fig:sup-mat-cosine-scissors}
\end{figure*}
        
\begin{figure*}[h]

	\begin{minipage}{.33\linewidth}
		\centering
		\subfloat[]{\label{fig:sup-mat-cosine-untrained-sink}\includegraphics[width=1\textwidth]{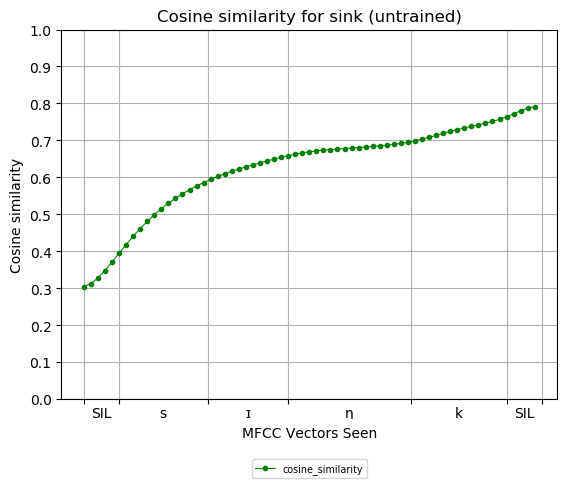}}
	\end{minipage}
	\hspace{-0.2cm}
	\begin{minipage}{.34\linewidth}
		\centering
		\subfloat[]{\label{fig:sup-mat-cosine-diff-sink}\includegraphics[width=1\textwidth]{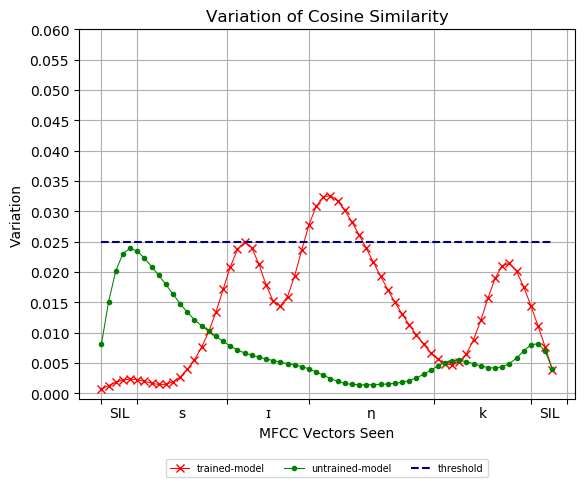}}
	\end{minipage}
	\hspace{-0.2cm}
	\begin{minipage}{.33\linewidth}
		\centering
		\subfloat[]{\label{fig:sup-mat-cosine-trained-sink}\includegraphics[width=1\textwidth]{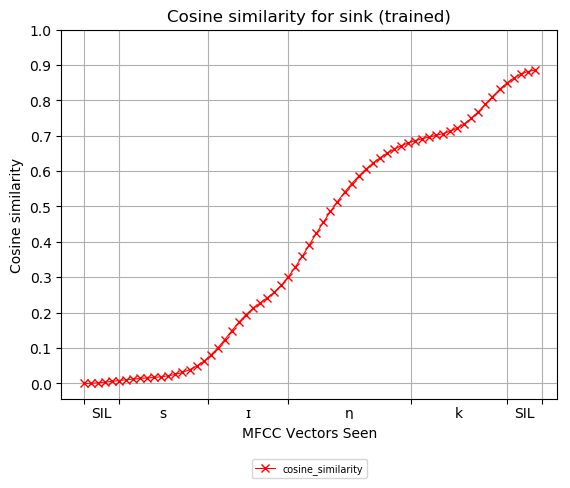}}
	\end{minipage}
	\caption{Evolution of cosine similarity for the word ``sink". Figure \ref{fig:sup-mat-cosine-diff-sink} shows peaks indicating the inflection points of curve \ref{fig:sup-mat-cosine-untrained-sink} (untrained model, green) and \ref{fig:sup-mat-cosine-trained-sink} (trained model, red)}
	\label{fig:sup-mat-cosine-sink}
\end{figure*}
        
\begin{figure*}[h]

	\begin{minipage}{.33\linewidth}
		\centering
		\subfloat[]{\label{fig:sup-mat-cosine-untrained-surfboard}\includegraphics[width=1\textwidth]{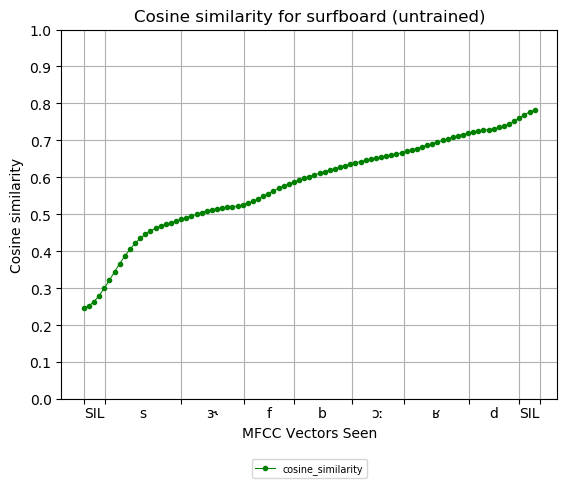}}
	\end{minipage}
	\hspace{-0.2cm}
	\begin{minipage}{.34\linewidth}
		\centering
		\subfloat[]{\label{fig:sup-mat-cosine-diff-surfboard}\includegraphics[width=1\textwidth]{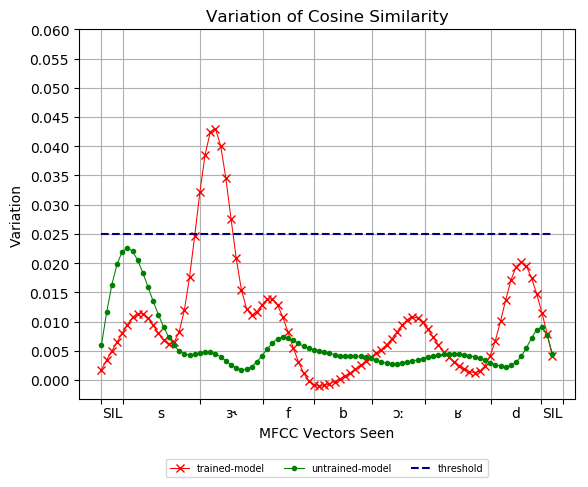}}
	\end{minipage}
	\hspace{-0.2cm}
	\begin{minipage}{.33\linewidth}
		\centering
		\subfloat[]{\label{fig:sup-mat-cosine-trained-surfboard}\includegraphics[width=1\textwidth]{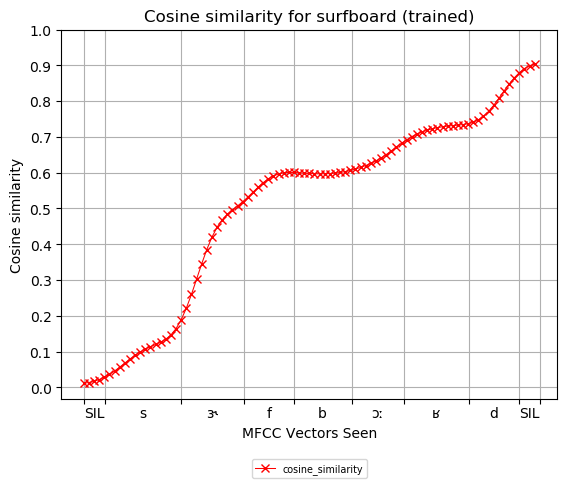}}
	\end{minipage}
	\caption{Evolution of cosine similarity for the word ``surfboard". Figure \ref{fig:sup-mat-cosine-diff-surfboard} shows peaks indicating the inflection points of curve \ref{fig:sup-mat-cosine-untrained-surfboard} (untrained model, green) and \ref{fig:sup-mat-cosine-trained-surfboard} (trained model, red)}
	\label{fig:sup-mat-cosine-surfboard}
\end{figure*}
        
\begin{figure*}[h]

	\begin{minipage}{.33\linewidth}
		\centering
		\subfloat[]{\label{fig:sup-mat-cosine-untrained-toilet}\includegraphics[width=1\textwidth]{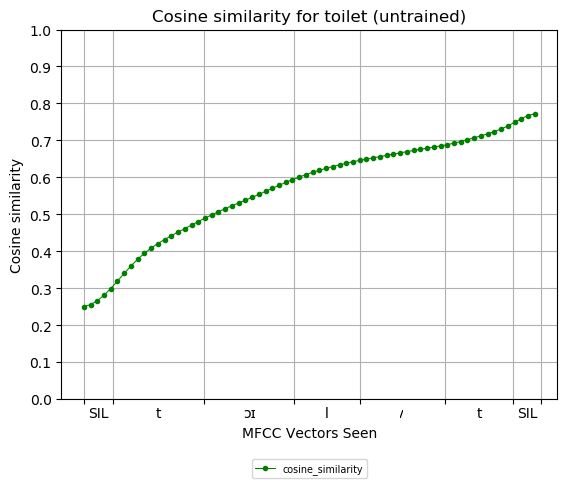}}
	\end{minipage}
	\hspace{-0.2cm}
	\begin{minipage}{.34\linewidth}
		\centering
		\subfloat[]{\label{fig:sup-mat-cosine-diff-toilet}\includegraphics[width=1\textwidth]{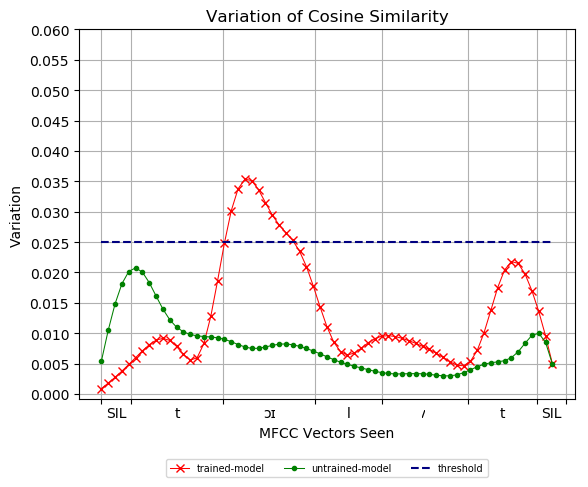}}
	\end{minipage}
	\hspace{-0.2cm}
	\begin{minipage}{.33\linewidth}
		\centering
		\subfloat[]{\label{fig:sup-mat-cosine-trained-toilet}\includegraphics[width=1\textwidth]{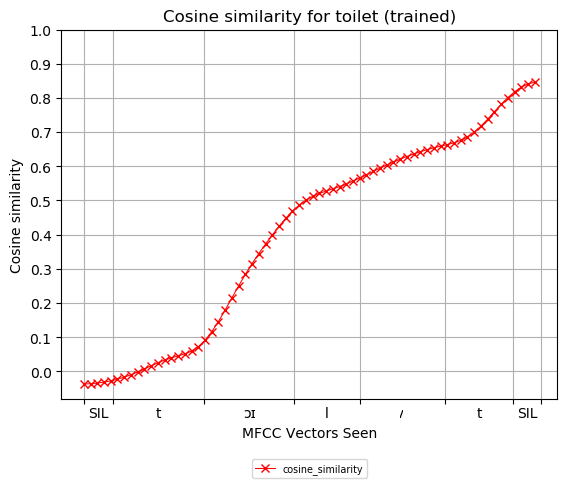}}
	\end{minipage}
	\caption{Evolution of cosine similarity for the word ``toilet". Figure \ref{fig:sup-mat-cosine-diff-toilet} shows peaks indicating the inflection points of curve \ref{fig:sup-mat-cosine-untrained-toilet} (untrained model, green) and \ref{fig:sup-mat-cosine-trained-toilet} (trained model, red)}
	\label{fig:sup-mat-cosine-toilet}
\end{figure*}
        
\begin{figure*}[h]

	\begin{minipage}{.33\linewidth}
		\centering
		\subfloat[]{\label{fig:sup-mat-cosine-untrained-traffic-light}\includegraphics[width=1\textwidth]{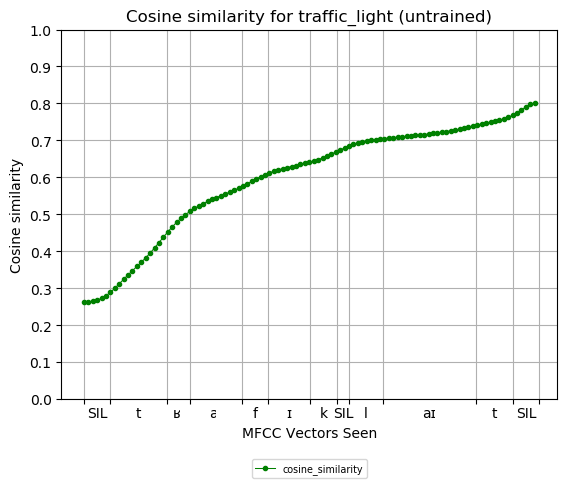}}
	\end{minipage}
	\hspace{-0.2cm}
	\begin{minipage}{.34\linewidth}
		\centering
		\subfloat[]{\label{fig:sup-mat-cosine-diff-traffic-light}\includegraphics[width=1\textwidth]{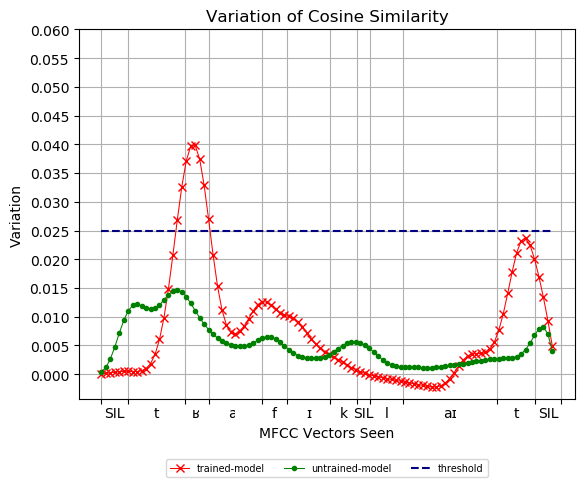}}
	\end{minipage}
	\hspace{-0.2cm}
	\begin{minipage}{.33\linewidth}
		\centering
		\subfloat[]{\label{fig:sup-mat-cosine-trained-traffic-light}\includegraphics[width=1\textwidth]{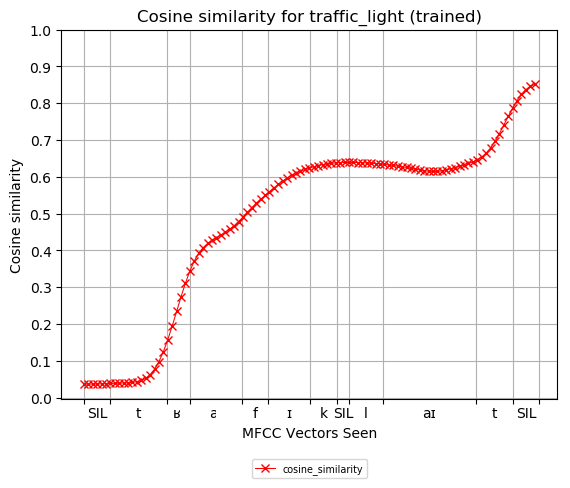}}
	\end{minipage}
	\caption{Evolution of cosine similarity for the word ``traffic light". Figure \ref{fig:sup-mat-cosine-diff-traffic-light} shows peaks indicating the inflection points of curve \ref{fig:sup-mat-cosine-untrained-traffic-light} (untrained model, green) and \ref{fig:sup-mat-cosine-trained-traffic-light} (trained model, red)}%
	\label{fig:sup-mat-cosine-fork}
\end{figure*}

\begin{figure*}[h]

	\begin{minipage}{.33\linewidth}
		\centering
		\subfloat[]{\label{fig:sup-mat-cosine-untrained-train}\includegraphics[width=1\textwidth]{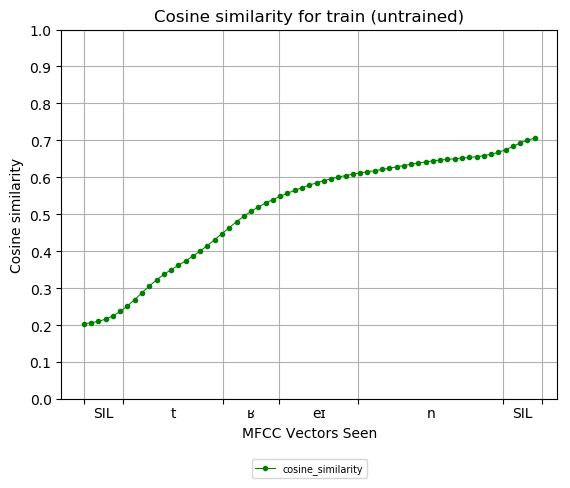}}
	\end{minipage}
	\hspace{-0.2cm}
	\begin{minipage}{.34\linewidth}
		\centering
		\subfloat[]{\label{fig:sup-mat-cosine-diff-train}\includegraphics[width=1\textwidth]{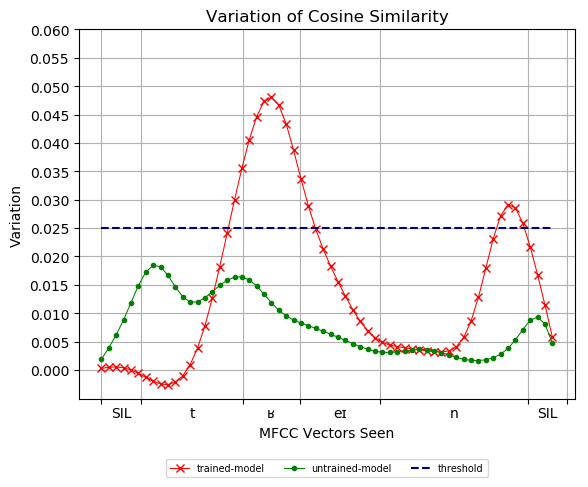}}
	\end{minipage}
	\hspace{-0.2cm}
	\begin{minipage}{.33\linewidth}
		\centering
		\subfloat[]{\label{fig:sup-mat-cosine-trained-train}\includegraphics[width=1\textwidth]{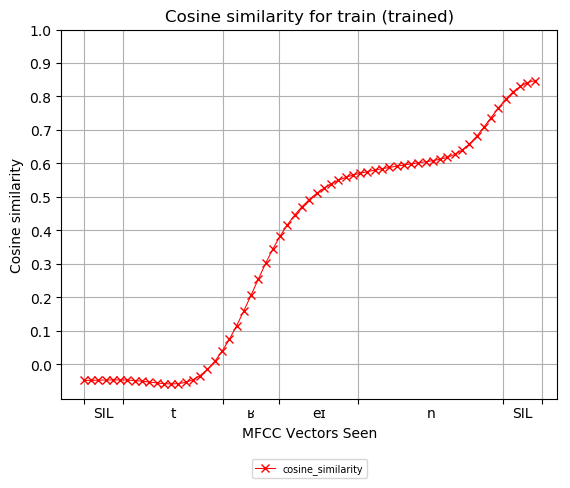}}
	\end{minipage}
	\caption{Evolution of cosine similarity for the word ``train". Figure \ref{fig:sup-mat-cosine-diff-train} shows peaks indicating the inflection points of curve \ref{fig:sup-mat-cosine-untrained-train} (untrained model, green) and \ref{fig:sup-mat-cosine-trained-train} (trained model, red)}
	\label{fig:sup-mat-cosine-train}
\end{figure*}

\begin{figure*}[h]

	\begin{minipage}{.33\linewidth}
		\centering
		\subfloat[]{\label{fig:sup-mat-cosine-untrained-wine-glass}\includegraphics[width=1\textwidth]{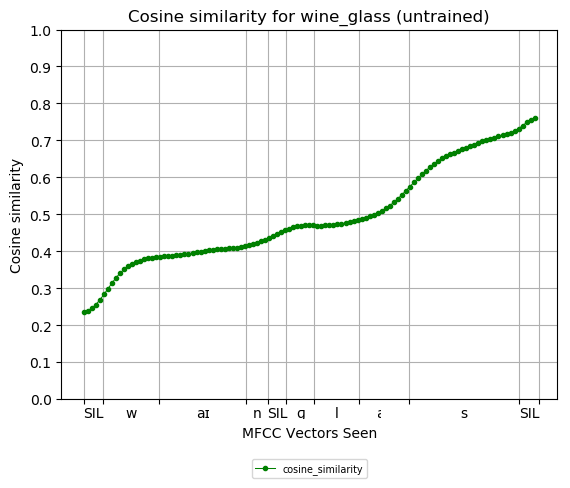}}
	\end{minipage}
	\hspace{-0.2cm}
	\begin{minipage}{.34\linewidth}
		\centering
		\subfloat[]{\label{fig:sup-mat-cosine-diff-wine-glass}\includegraphics[width=1\textwidth]{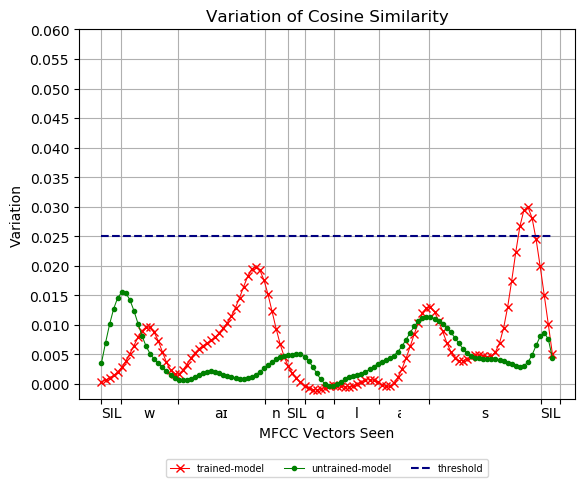}}
	\end{minipage}
	\hspace{-0.2cm}
	\begin{minipage}{.33\linewidth}
		\centering
		\subfloat[]{\label{fig:sup-mat-cosine-trained-wine-glass}\includegraphics[width=1\textwidth]{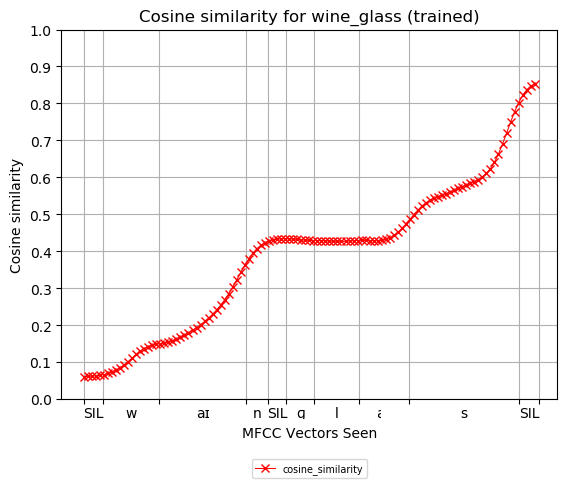}}
	\end{minipage}
	\caption{Evolution of cosine similarity for the word ``wine glass". Figure \ref{fig:sup-mat-cosine-diff-wine-glass} shows peaks indicating the inflection points of curve \ref{fig:sup-mat-cosine-untrained-wine-glass} (untrained model, green) and \ref{fig:sup-mat-cosine-trained-wine-glass} (trained model, red)}
	\label{fig:sup-mat-cosine-wine_glass}
\end{figure*}

\begin{figure*}[h]

	\begin{minipage}{.33\linewidth}
		\centering
		\subfloat[]{\label{fig:sup-mat-cosine-untrained-zebra}\includegraphics[width=1\textwidth]{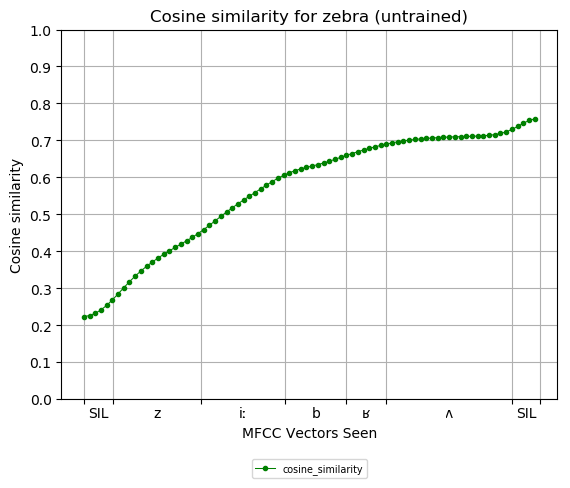}}
	\end{minipage}
	\hspace{-0.2cm}
	\begin{minipage}{.34\linewidth}
		\centering
		\subfloat[]{\label{fig:sup-mat-cosine-diff-zebra}\includegraphics[width=1\textwidth]{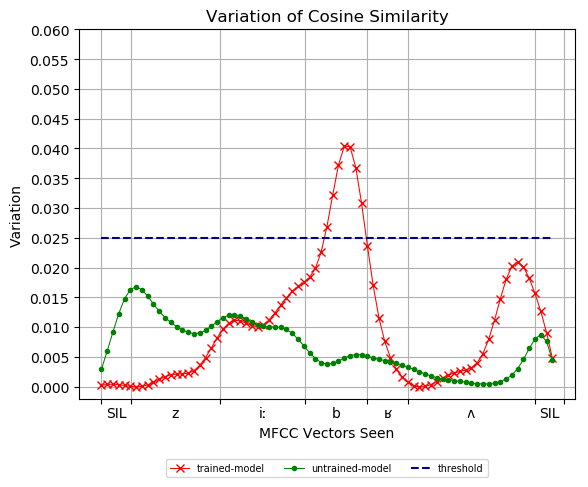}}
	\end{minipage}
	\hspace{-0.2cm}
	\begin{minipage}{.33\linewidth}
		\centering
		\subfloat[]{\label{fig:sup-mat-cosine-trained-zebra}\includegraphics[width=1\textwidth]{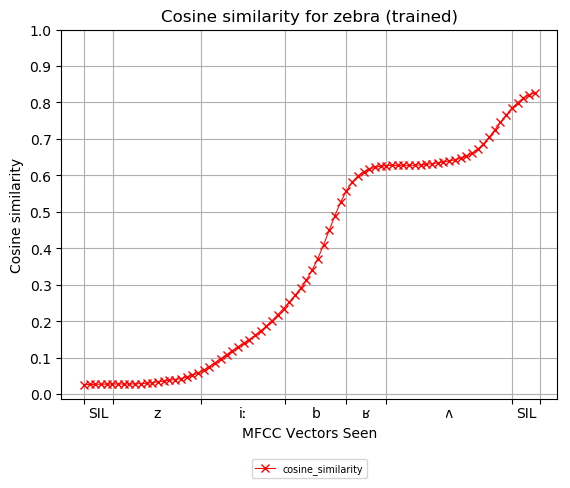}}
	\end{minipage}
	\caption{Evolution of cosine similarity for the word ``zebra". Figure \ref{fig:sup-mat-cosine-diff-zebra} shows peaks indicating the inflection points of curve \ref{fig:sup-mat-cosine-untrained-zebra} (untrained model, green) and \ref{fig:sup-mat-cosine-trained-zebra} (trained model, red)}
	\label{fig:sup-mat-cosine-zebra}
\end{figure*}

\begin{figure*}[h]
\subsection{Competition Sample}
\label{sec:supplementary-competition}
    \begin{minipage}{.50\linewidth}
		\centering
		\subfloat[]{\label{fig:competition_baseball-baby}\includegraphics[width=1\textwidth]{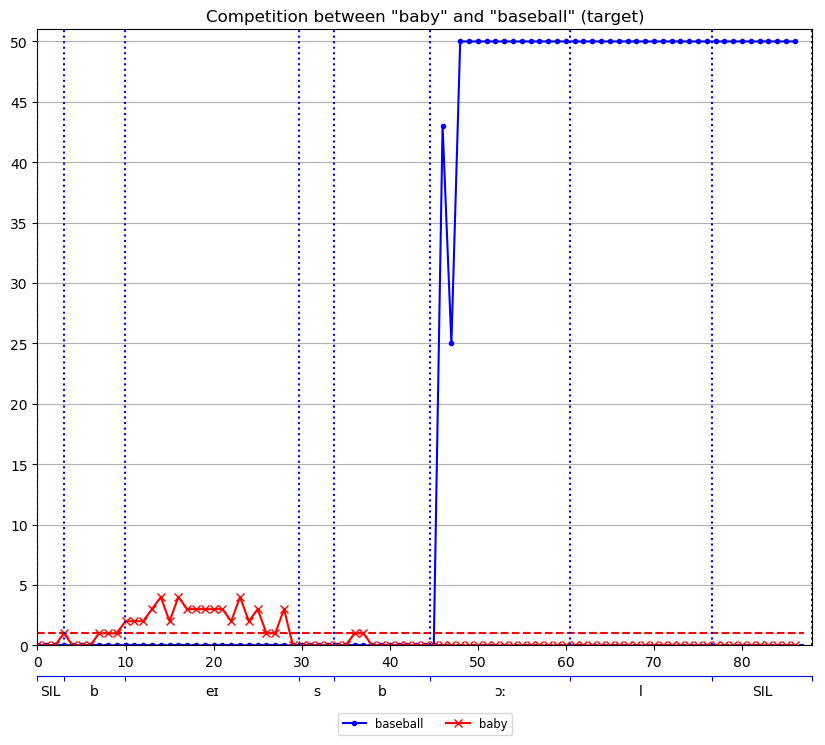}}
	\end{minipage}
	\begin{minipage}{.50\linewidth}
		\centering
		\subfloat[]{\label{fig:competition_bed-bench}\includegraphics[width=1\textwidth]{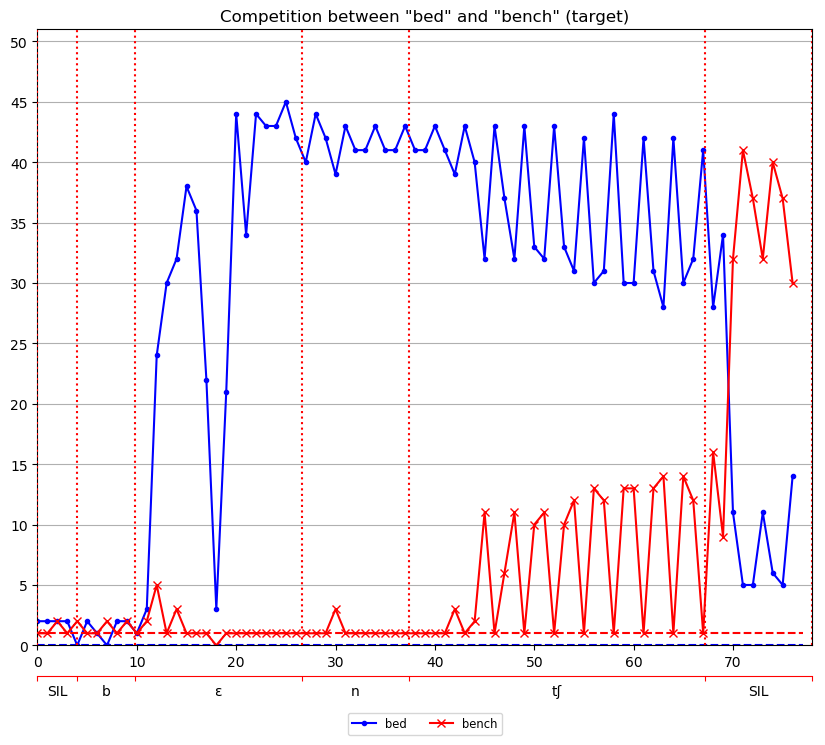}}
	\end{minipage}
	\caption{Competition plots between \ref{fig:competition_baseball-baby} ``baseball" and ``baby" and \ref{fig:competition_bed-bench} ``bed" and ``bench".}
\end{figure*}

\begin{figure*}[h]
    \begin{minipage}{.50\linewidth}
		\centering
		\subfloat[]{\label{fig:competition_glass-grass}\includegraphics[width=1\textwidth]{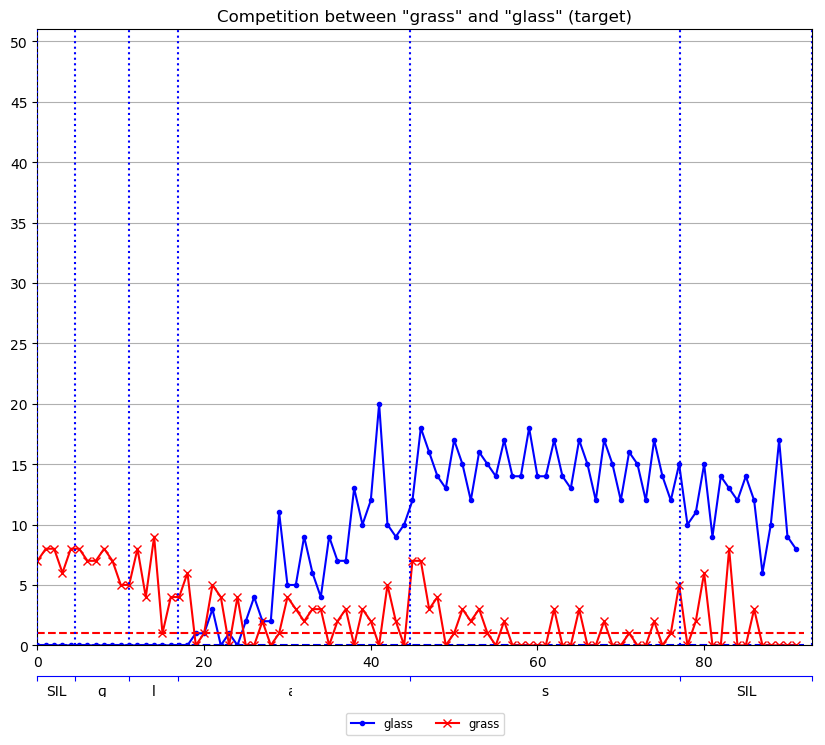}}
	\end{minipage}
	\begin{minipage}{.50\linewidth}
		\centering
		\subfloat[]{\label{fig:competition_kitchen-keyboard}\includegraphics[width=1\textwidth]{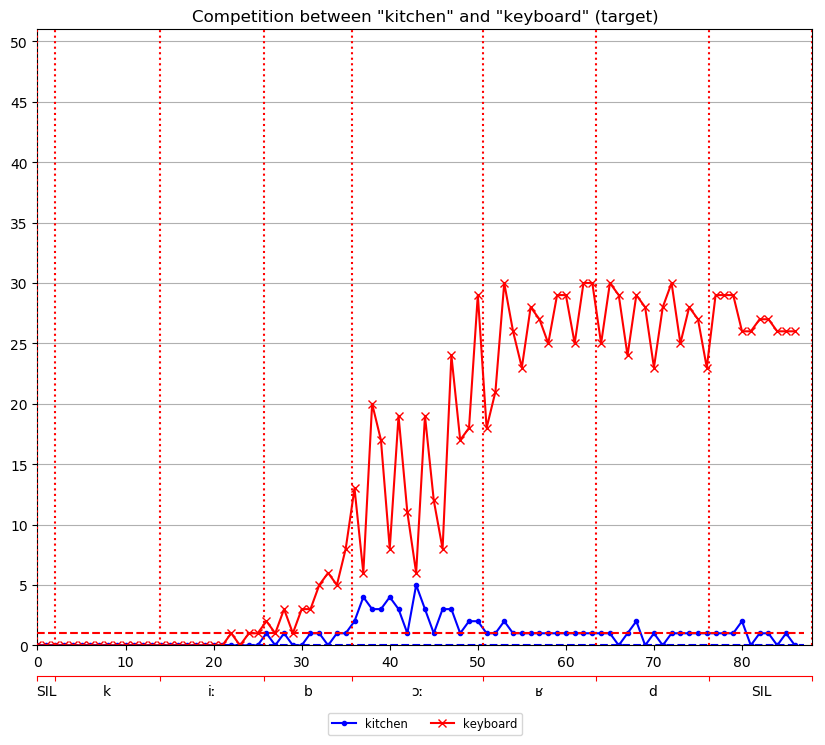}}
	\end{minipage}
	\caption{Competition plots between \ref{fig:competition_glass-grass} ``glass" and ``grass" and \ref{fig:competition_kitchen-keyboard} ``kitchen" and ``keyboard".}
\end{figure*}

\begin{figure*}[h]
    \begin{minipage}{.50\linewidth}
		\centering
		\subfloat[]{\label{fig:competition_meat-mirror}\includegraphics[width=1\textwidth]{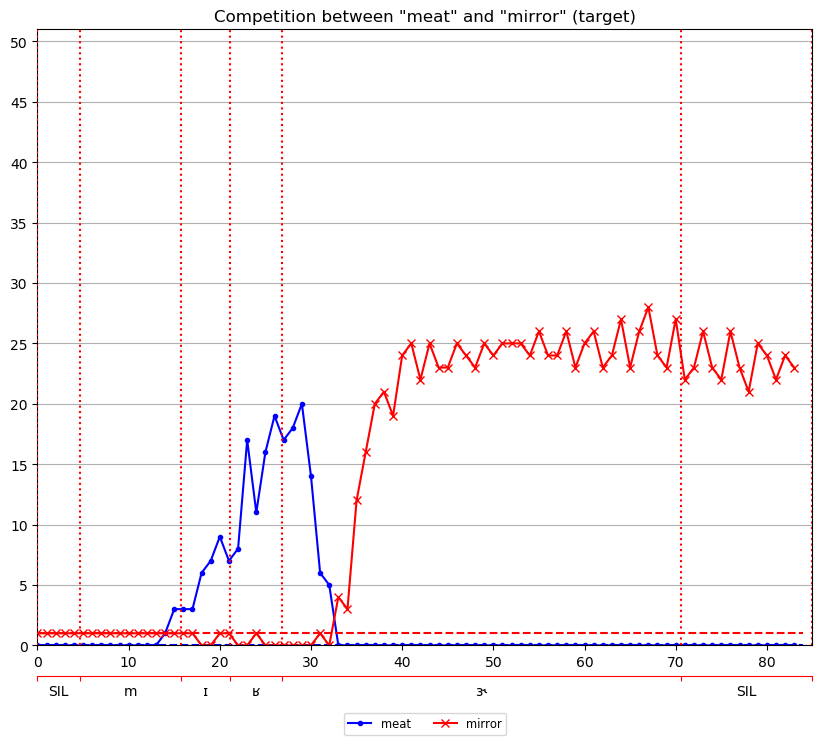}}
	\end{minipage}
	\begin{minipage}{.50\linewidth}
		\centering
		\subfloat[]{\label{fig:competition_mirror-meter}\includegraphics[width=1\textwidth]{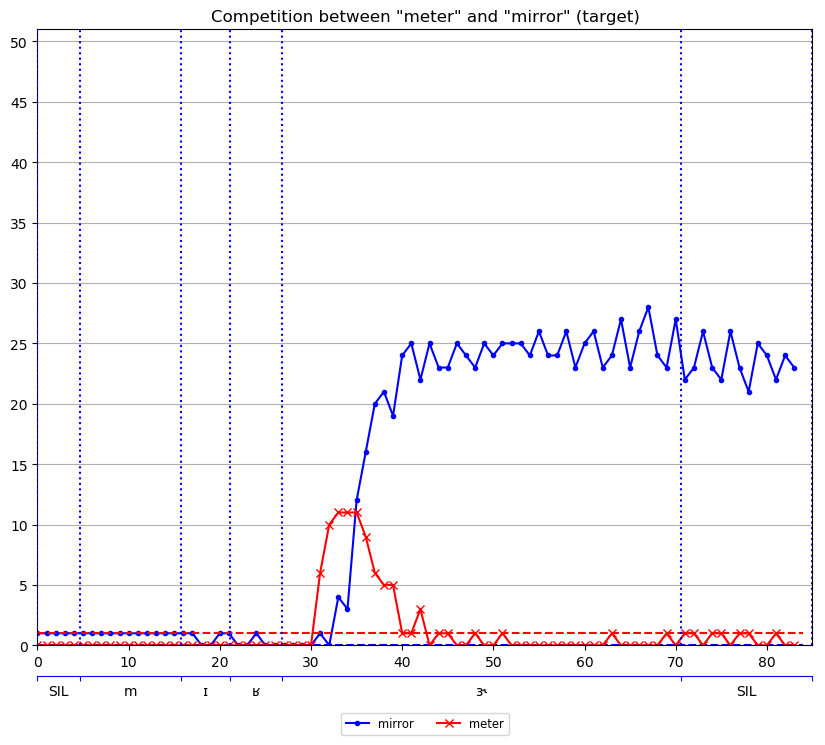}}
	\end{minipage}
	\caption{Competition plots between \ref{fig:competition_meat-mirror} ``meat" and ``mirror" and \ref{fig:competition_mirror-meter} ``mirror" and ``meter".}
\end{figure*}

\begin{figure*}[h]
    \begin{minipage}{.50\linewidth}
		\centering
		\subfloat[]{\label{fig:competition_plane-plate}\includegraphics[width=1\textwidth]{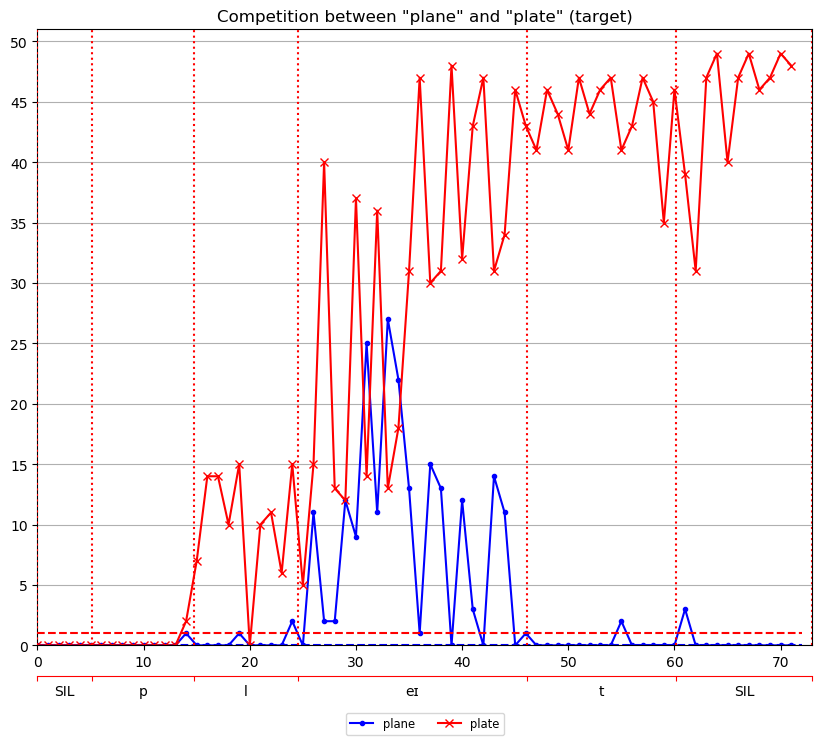}}
	\end{minipage}
	\begin{minipage}{.50\linewidth}
		\centering
		\subfloat[]{\label{fig:competition_shore-shower}\includegraphics[width=1\textwidth]{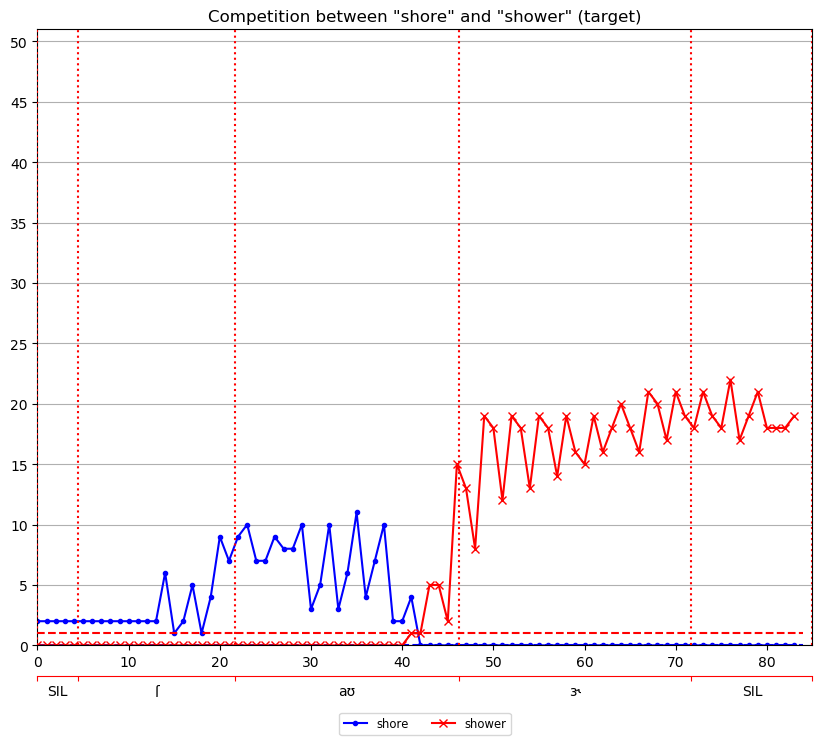}}
	\end{minipage}
	\caption{Competition plots between \ref{fig:competition_plane-plate} ``plane" and ``plate" and \ref{fig:competition_shore-shower} ``shore" and ``shower".}
\end{figure*}

\end{document}